%% file: main_arxiv.tex
\documentclass{article} 
\usepackage{iclr2027_conference,times}

\input{math_commands.tex}

\usepackage{hyperref}
\usepackage{url}
\usepackage{graphicx}
\newcommand{\libplus}{{LIBERO-Plus}}
\newcommand{\bench}{\textbf{{VLA-Retain}}}
 \usepackage{booktabs} 
\usepackage{tabularx}
\usepackage{pifont}

\usepackage{tcolorbox}
\definecolor{DraftAmber}{RGB}{145,77,0}
\definecolor{SlotBackground}{RGB}{245,247,249}
\definecolor{SlotBorder}{RGB}{158,172,185}
\definecolor{FindingBlue}{RGB}{27,58,91}
\definecolor{FindingBackground}{RGB}{242,247,251}
\definecolor{MethodRowGray}{RGB}{238,238,238}
\newtcolorbox{finding}[2]{colback=FindingBackground,colframe=FindingBlue,
  colbacktitle=FindingBlue,coltitle=white,fonttitle=\bfseries,
  title={Finding #1: #2},arc=0pt,boxrule=0.5pt,boxsep=0pt,
  left=6pt,right=6pt,top=5pt,bottom=5pt,toptitle=4pt,bottomtitle=4pt,
  before skip=7pt,after skip=7pt}

\title{VLA-Retain: Benchmarking Cross-Dimension Effects of Vision Tuning in VLAs}

\author{
Borui Xu\textsuperscript{1},
Han Chen\textsuperscript{1},
Bingsheng He\textsuperscript{1},
Ming Wu\textsuperscript{2},
Jason Zeng\textsuperscript{2},
Michael Heinrich\textsuperscript{2} \\
\textsuperscript{1}National University of Singapore \\
\textsuperscript{2}0G Labs
}

\iclrfinalcopy 
\begin{document}

\maketitle

\begin{abstract}

Vision-language-action (VLA) models transfer broad visual and semantic knowledge from large-scale pretraining into robot control, but their manipulation performance remains sensitive to changes in observation conditions, which recent work addresses by fine-tuning the visual module on target-domain data. However, it remains unclear how such adaptation affects robustness beyond the target condition, as existing robustness benchmarks only evaluate models without fine-tuning and adaptation studies only report the target gain. Measuring this cross-dimension effect requires separating target gains from off-target changes, isolating adaptation effects from architectural differences, ruling out generic forgetting, and comparing tuning choices. Therefore, we introduce \bench{}, a paired benchmark that compares five open VLA models, tuned on Camera-view or Sensor-noise data, with their own base checkpoints on the same 10,030 \libplus{} episodes spanning seven perturbation dimensions under varied training data and tuning strategies. Our study reveals four findings. First, targeted vision tuning is not local, and its target gain does not predict retention. Mean off-target gain is negative in 8 of 10 model--domain configurations. Under Camera-view tuning, X-VLA gains $69.4$ percentage points on the target but loses $4.6$ points on average elsewhere, whereas OpenVLA-OFT gains $39.3$ and $1.4$ points, respectively. Second, these regressions are not explained only by performance loss on the same tasks without perturbations. Third, broader training coverage raises overall gain over target-specific training but does not eliminate dimension-level regressions. Fourth, tuning strategies are model-dependent, as DELTA-style feature preservation raises X-VLA's mean off-target gain to $+1.2$ points but lowers OpenVLA-OFT's to $-23.1$. Overall, these findings show that target-domain success alone is insufficient and highlight cross-dimensional retention as a key criterion for targeted VLA adaptation.

\end{abstract}

\input{01-introduction}
\input{02-background}
\input{03-design}

\input{04-evaluation}

\input{05-conclusion}

\bibliography{iclr2027_conference}
\bibliographystyle{iclr2027_conference}

\appendix
\input{appendix}

\end{document}

%% file: math_commands.tex
\usepackage{amsmath,amsfonts,bm}

\def\eqref#1{equation~\ref{#1}}

\def\1{\bm{1}}

\DeclareMathAlphabet{\mathsfit}{\encodingdefault}{\sfdefault}{m}{sl}
\SetMathAlphabet{\mathsfit}{bold}{\encodingdefault}{\sfdefault}{bx}{n}



%% file: 01-introduction.tex
\section{Introduction}

\label{sec:introduction}

Vision-language-action (VLA) models map visual observations, language instructions, and robot state to executable actions, transferring large-scale visual and semantic knowledge into robot control~\citep{brohan2023rt2,driess2023palme,openx2023,kim2024openvla}. This provides a scalable route toward general-purpose manipulation, but deployment conditions rarely match those seen during training, as camera placement changes, sensors become noisy, illumination and backgrounds vary, objects appear in new layouts, and operators phrase the same goal differently. Current VLAs remain sensitive to these shifts~\citep{simplerenv2024,fei2025liberoplus}, so a deployed model must be adapted to new conditions. Reliable deployment therefore requires adapting to the new condition while retaining robustness to the other conditions the robot continues to encounter.

Existing studies address this requirement from two directions. First, robustness benchmarks~\citep{fei2025liberoplus,zhang2025vlaarena,wang2026liberox} evaluate fixed models under controlled perturbations, showing that robustness varies across visual, spatial, linguistic, and task-level dimensions and thereby guiding model selection. Second, targeted adaptation studies fine-tune the chosen model's vision encoder on target-domain data to recover robustness to a selected shift. \cite{li2025generalizable} recover camera-view robustness through low-rank visual updates, while VLAGuard~\citep{yin2026vlaguard} and Structure-Aware Robust Fine-Tuning~\citep{zhang2026sarf} defend against physical attention hijacking by tuning only the vision encoder. Since model selection cannot cover every shift, targeted vision tuning offers a more practical adaptation strategy.

However, this strategy overlooks how tuning for one shift affects robustness to non-target perturbations. Targeted adaptation studies report only the target-domain gain, and robustness benchmarks evaluate only models without such tuning, so neither reveals \textbf{how the adaptation itself changes robustness under the remaining conditions}. This matters because the adapted visual representation feeds the same frozen downstream components under every condition. Tuning it for one domain can thus change behavior under unseen shifts in language, object layout, or sensor noise, while target-domain success or an aggregate score can hide such regressions.

Measuring such cross-dimension effects poses four challenges. First, adaptation effects are multi-dimensional, so neither target-only nor aggregate evaluation suffices. Second, VLA architectures differ in baseline robustness, so absolute post-tuning scores mix adaptation effects with differences in their starting points. Third, an off-target regression may reflect generic forgetting or task sampling rather than perturbation-specific interference. Fourth, adaptation has a large design space spanning data coverage, tuning scope, and retention methods, whose effects may vary by model.

\begin{figure}[]
  \centering
  \includegraphics[width=0.93\linewidth]{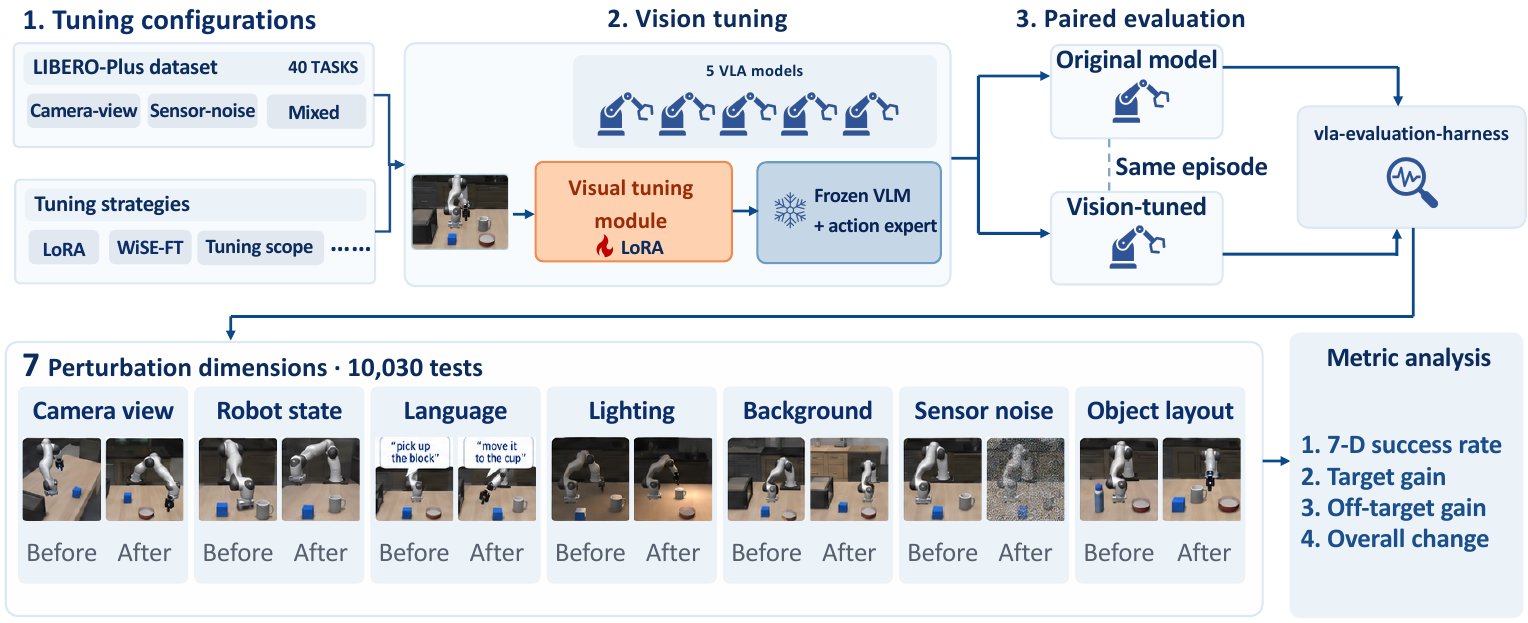}
  \vspace{-1em}
 \caption{\textbf{Overview of \bench{}.} Target-domain data update only the visual module, but the updated module processes observations under every deployment condition. \bench{} therefore evaluates each base--tuned pair on the same episodes across seven perturbation dimensions and reports both target improvement and off-target retention.}
  \label{fig:benchmark_overview}
  \vspace{-1.3em}
\end{figure}

We introduce \bench{}, a paired benchmark designed to address these challenges (Figure~\ref{fig:benchmark_overview}). To expose multi-dimensional effects, it evaluates every base--tuned pair on the same 10,030 \libplus{} episodes~\citep{fei2025liberoplus} including seven perturbation dimensions and reports target, mean and worst off-target, and overall gains separately. To isolate adaptation effects, it compares five open VLA models, each tuned on Camera-view or Sensor-noise data, with their own base checkpoints fine-tuned on LIBERO~\citep{liu2023libero}. To test alternative explanations, it adds original-LIBERO controls and paired cluster-bootstrap intervals. To cover the design space, it varies training-data coverage across all five models and tuning strategy on two representative architectures. We use \bench{} to answer the following four questions.

\textbf{Does targeted vision tuning affect off-target robustness?} Yes, and the change is usually a regression. Mean off-target gain is negative in 8 of 10 model--domain configurations, and language and object-layout success decline in 9 of 10, although only the visual module is updated. The target gain does not predict these effects: under Camera-view tuning, X-VLA improves target success from $19.6\%$ to $89.0\%$ but loses $4.6$ percentage points on average across off-target dimensions, whereas OpenVLA-OFT improves from $55.2\%$ to $94.5\%$ while gaining $1.4$ points.

\textbf{Are these regressions explained by forgetting of the original tasks?} Not fully. Success on the same tasks without perturbations declines by at most $3.8$ points for every checkpoint except NORA-1.5 after Camera-view tuning. After subtracting this change, every Camera-view-tuned model and four of five Sensor-noise-tuned models still regress on at least one off-target perturbation.

\textbf{Does broader training coverage prevent them?} No. Mixed training over the complete \libplus{} training set gets a higher overall gain than either target-specific setting for all five models, yet every model still regresses on at least one dimension.

\textbf{Can alternative tuning strategies control them?} Not consistently across models. Weight interpolation improves retention on both evaluated models but reduces the target gain, and the preferred tuning scope differs between them. In particular, DELTA-style feature preservation raises X-VLA's mean off-target gain from $-4.6$ to $+1.2$ points but lowers OpenVLA-OFT's from $+1.4$ to $-23.1$.

These results show that target-domain evaluation is valid but incomplete, as the same update can also affect off-target performance. By treating each tuned model as a paired change across seven robustness dimensions, \bench{} separates target gains from changes elsewhere and makes cross-dimensional retention measurable. Targeted VLA adaptation should therefore be evaluated jointly by target improvement, off-target retention, and overall performance.

%% file: 02-background.tex
\section{Related Work}
\label{sec:background}



\textbf{Targeted adaptation of VLA models.} VLA models map images and language instructions, optionally together with robot proprioception, to discrete action tokens or continuous action chunks~\citep{brohan2023rt2,driess2023palme,openx2023}. Modern systems differ in their visual backbones, multimodal trunks, and action decoders~\citep{kim2024openvla,kim2025oft,black2025pi05,zheng2025xvla,hung2025nora,ma2026internvla}, but all rely on a visual representation that is consumed by downstream reasoning and action components. Parameter-efficient adaptation, including LoRA~\citep{hu2022lora}, makes it practical to update this pathway while freezing most of the VLA. Recent robustness methods use such visual updates to correct spatial misalignment under camera changes or defend against physical attention hijacking~\citep{li2025generalizable,yin2026vlaguard,zhang2026sarf}, establishing that adaptation improves the target condition. \bench{} does not challenge these target-domain gains, it asks what the tuning update changes elsewhere.



\textbf{Retention under model adaptation.} Retention methods constrain adaptation by keeping the tuned model close to a pretrained solution. WiSE-FT interpolates pretrained and fine-tuned weights to improve robustness~\citep{wortsman2022robust}, while DELTA preserves task-relevant source representations during transfer~\citep{li2018delta}. Our setting differs from sequential task learning because the manipulation tasks remain fixed while adaptation targets a specific observation shift in the shared visual pathway. We therefore evaluate retention across deployment conditions rather than by accuracy over a sequence of learned tasks, and Section~\ref{sec:mitigation} tests whether such retention methods control cross-dimension effects in VLAs.

\textbf{VLA robustness benchmarks.} Existing benchmarks expose complementary limitations of VLA models. SimplerEnv~\citep{simplerenv2024} studies simulation-based evaluation for real-robot setups, while LIBERO-Plus~\citep{fei2025liberoplus} extends LIBERO's~\citep{liu2023libero} manipulation tasks with seven controlled perturbation dimensions. VLA-Arena~\citep{zhang2025vlaarena} organizes evaluation by task structure, language, and visual difficulty, and LIBERO-X~\citep{wang2026liberox} introduces hierarchical tests of spatial generalization, object recognition, and instruction understanding. VLM4VLA~\citep{zhang2026vlm4vla} uses a common action design to isolate how VLM choice and embodied capabilities affect downstream control, identifying the visual module as a central bottleneck. These works characterize the robustness of fixed models, backbones, or generalization regimes, whereas \bench{} measures how perturbation-specific vision tuning changes that robustness by pairing every fine-tuned model with its own base model.


%% file: 03-design.tex
\section{Studying Cross-Dimension Effects of VLA Adaptation}
\label{sec:evaluation_view}

\subsection{Research Problem}
\label{sec:research_problem}

We view a VLA as a visual module followed by downstream multimodal and action components:
\begin{equation}
a=g_{\phi}\!\left(f_{\theta}(x),\ell,s\right),
\label{eq:vla_decomposition}
\end{equation}
where $f_{\theta}$ denotes the visual backbone and connector, $g_{\phi}$ denotes the remaining VLA, and $(x,\ell,s)$ are the image observation, language instruction, and robot state. Target-specific vision tuning changes $\theta$ to $\theta'$ while keeping $\phi$ fixed. The parameter update is constrained to the visual module, but the adapted module processes observations under every perturbation condition.

We therefore describe each model by its robustness profile over the perturbation dimensions $\mathcal D$. Let $S^{\mathrm{base}}_{m,d}$ denote the success rate, expressed as a percentage, of base model $m$ on perturbation dimension $d$, and let $S^{\mathrm{tuned}}_{m,c,d}$ denote the corresponding success rate after applying tuning configuration $c$. The object of study is the change vector $\boldsymbol{\Delta}_{m,c}=(\Delta_{m,c,d})_{d\in\mathcal D}$ induced by tuning rather than either profile alone, with entries measured in percentage points (pp):
\begin{equation}
\Delta_{m,c,d}=S^{\mathrm{tuned}}_{m,c,d}-S^{\mathrm{base}}_{m,d}.
\label{eq:dimension_change}
\end{equation}
For a target dimension $t$, $\Delta_{m,c,t}$ measures the intended improvement, while the other six values measure how tuning affects robustness outside the training domain. Negative values indicate regression, and values near zero indicate retention. Robustness benchmarks characterize $S^{\mathrm{base}}_{m,d}$ for fixed models, and targeted adaptation studies report $\Delta_{m,c,t}$; we study the complete vector $\boldsymbol{\Delta}_{m,c}$.

\subsection{Design Principles}
\label{sec:design_principles}

\bench{} follows four design principles, each addressing one of the challenges in Section~\ref{sec:introduction}.

\textbf{Full-profile evaluation.} A large target gain can dominate an aggregate score, and a positive mean off-target change can still hide a severe drop on one dimension. We therefore evaluate every tuned model on all seven dimensions and summarize the change vector with target, mean off-target, worst off-target, and episode-weighted overall gains:
\begin{equation}
\begin{aligned}
G^{\mathrm{target}}_{m,c}&=\Delta_{m,c,t}, & G^{\mathrm{off}}_{m,c}&=\frac{1}{|\mathcal D|-1}\sum_{d\neq t}\Delta_{m,c,d},\\
G^{\mathrm{worst}}_{m,c}&=\min_{d\neq t}\Delta_{m,c,d}, & G^{\mathrm{all}}_{m,c}&=\frac{\sum_{d\in\mathcal D}N_d\Delta_{m,c,d}}{\sum_{d\in\mathcal D}N_d},
\end{aligned}
\label{eq:retention_metrics}
\end{equation}
where $N_d$ is the number of episodes in dimension $d$; all four metrics are in percentage points.

\textbf{Paired comparison.} Absolute tuned scores conflate adaptation with baseline robustness, so we pair every tuned checkpoint with its own base checkpoint on identical episodes.

\textbf{Controls for alternative explanations.} To separate perturbation-specific regression from generic forgetting, task sampling, checkpoint selection, and training randomness, we add original-LIBERO controls, paired cluster-bootstrap intervals, and checkpoint and seed analyses (Appendix~\ref{app:checkpoint_analysis}).

\textbf{Varied adaptation strategies.} Because a conclusion drawn from one tuning strategy may not hold for others, we vary training-data coverage across all five models and compare tuning scope, weight interpolation, and feature preservation on two representative architectures.

\section{\bench{} Benchmark Design}
\label{sec:benchmark}

\subsection{Model Coverage}

To address architectural heterogeneity, we study five open-source VLA models spanning different visual interfaces and action-generation mechanisms. Every base model is the released checkpoint after LIBERO fine-tuning rather than a generic pretrained checkpoint. Table~\ref{tab:model_summary} summarizes the main architectural differences. We define the visual module as the visual backbone together with the projector or merger that maps visual features into the multimodal representation space. Appendix Figure~\ref{fig:architecture_comparison} and Table~\ref{tab:model_parameters} provide the complete visual interfaces and parameter counts.

\begin{table}[]
  \caption{Architectural coverage of the core benchmark. Visual share denotes the fraction of total parameters contained in the visual backbone and connector.}
  \label{tab:model_summary}
  \centering
  \setlength{\tabcolsep}{2pt}
  \renewcommand{\arraystretch}{1.02}
  \begin{tabularx}{\linewidth}{@{}l>{\raggedright\arraybackslash}p{0.20\linewidth}>{\raggedright\arraybackslash}X>{\raggedleft\arraybackslash}p{0.09\linewidth}@{}}
    \toprule
    \textbf{Model} & \textbf{Visual backbone} & \textbf{Action formulation} & \multicolumn{1}{c}{\shortstack{\textbf{Visual}\\\textbf{share}}} \\
    \midrule
    OpenVLA-OFT~\citep{kim2025oft} & SigLIP + DINOv2 & Regression head & $10.64\%$ \\
    $\pi_{0.5}$~\citep{black2025pi05} & SigLIP & Flow matching & $11.40\%$ \\
    InternVLA-A1.5~\citep{ma2026internvla} & Qwen3.5 ViT & Flow matching & $12.34\%$ \\
    NORA-1.5~\citep{hung2025nora} & Qwen2.5-VL ViT & Flow matching + FAST & $16.77\%$ \\
    X-VLA~\citep{zheng2025xvla} & DaViT & \mbox{Soft-prompted flow matching} & $41.26\%$ \\
    \bottomrule
  \end{tabularx}

  \vspace{-0.4em}
\end{table}

\subsection{Training Data and Evaluation Dimensions}

The main targeted settings use the official \libplus{}~\citep{fei2025liberoplus} Camera-view and Sensor-noise training subsets, which represent changes in viewpoint and low-level image noise. Each subset isolates one target perturbation while covering the same 40 base manipulation tasks. To study training-data coverage, we additionally use Mixed training, which draws from the complete \libplus{} training dataset and covers five of the seven evaluation dimensions, all except robot state and object layout. It therefore has no single target dimension. All three settings use the same evaluation set and optimization budget; they differ in the perturbation coverage of the training data.



To expose multi-dimensional effects, evaluation uses a separate \libplus{} evaluation set of 10,030 perturbation-specific episodes derived from the same 40 source tasks and spanning the seven dimensions in Table~\ref{tab:evaluation_set}. The training and evaluation sets share task definitions but contain different perturbation instances. Training data determine which perturbations are observed during tuning, whereas every base--tuned pair is evaluated on the complete set with identical episode keys.


\begin{table}[t]
  \caption{Evaluation composition across perturbation dimensions and LIBERO task suites.}
  \label{tab:evaluation_set}
  \centering
  \setlength{\tabcolsep}{6pt}
  \renewcommand{\arraystretch}{1.02}
  \begin{tabular*}{\linewidth}{@{\extracolsep{\fill}}lrrrrr@{}}
    \toprule
    \textbf{Dimension} & \textbf{Spatial} & \textbf{Object} & \textbf{Goal} & \textbf{LIBERO-10} & \textbf{Total} \\
    \midrule
    Camera view & 376 & 396 & 408 & 419 & 1,599 \\
    Robot state & 350 & 398 & 409 & 393 & 1,550 \\
    Language & 390 & 354 & 410 & 383 & 1,537 \\
    Lighting & 292 & 297 & 279 & 274 & 1,142 \\
    Background & 258 & 248 & 281 & 289 & 1,076 \\
    Sensor noise & 351 & 422 & 379 & 449 & 1,601 \\
    Object layout & 385 & 403 & 425 & 312 & 1,525 \\
    \midrule
    \textbf{Total} & \textbf{2,402} & \textbf{2,518} & \textbf{2,591} & \textbf{2,519} & \textbf{10,030} \\
    \bottomrule
  \end{tabular*}
  \vspace{-0.4em}
\end{table}

\subsection{Tuning Configurations and Training Protocol}

The default training setting is Dual LoRA, which applies rank-32 LoRA to both the visual backbone and connector while freezing the multimodal backbone and action module. This reference configuration is used for the five-model comparisons under Camera-view, Sensor-noise, and Mixed training. Each model retains its official learning-rate schedule and normalization statistics.

To study the adaptation design space, we compare the tuning configurations in Appendix Table~\ref{tab:tuning_strategies} on X-VLA and OpenVLA-OFT using Camera-view training data. The tuning scope variants change whether LoRA is applied to the visual backbone, the connector, or both, while keeping the downstream modules frozen. WiSE-FT~\citep{wortsman2022robust} interpolates the visual weights of the base and Dual-LoRA models without additional training. Based on DELTA~\citep{li2018delta}, DELTA-style starts from Dual LoRA and adds a Taylor-weighted penalty that anchors the tuned post-connector representation to a frozen base branch; Appendix~\ref{app:delta_style} gives the complete objective. Each comparison otherwise uses the same model, training data, optimization budget, and evaluation episodes.

All training-based configurations use AdamW~\citep{loshchilov2019decoupled} in BF16 with an effective batch size of 32 for 12,000 optimizer updates. LoRA configurations use rank 32, $\alpha=16$, and zero dropout. Appendix Table~\ref{tab:training_hyperparameters} reports model-specific learning rates and schedules.


\subsection{Simulator and Paired Evaluation}

All rollouts are executed in the LIBERO simulator~\citep{liu2023libero} with \libplus{} perturbations~\citep{fei2025liberoplus}, using robosuite 1.4.1~\citep{robosuite2020} and MuJoCo physics~\citep{todorov2012mujoco}. We evaluate every base and tuned model with \texttt{vla-evaluation-harness}~\citep{choi2026vlaeval}, which separates model inference from simulator execution through a common interface. Model adapters preserve each model's native image preprocessing, proprioceptive inputs, normalization, action space, action chunks, gripper convention, and execution horizon. An episode is successful when the simulator's task-completion condition is reached. Because each base--tuned pair uses identical episode keys, every reported change is paired at the episode level.

All summary metrics follow Equation~\ref{eq:retention_metrics}. Mixed training has no unique target, so we report its seven dimension-level changes and overall gain without target or off-target aggregates. 


To quantify task-sampling uncertainty, we compute pointwise $95\%$ confidence intervals with a paired cluster bootstrap that resamples the 40 source tasks 20,000 times (Appendix~\ref{app:confidence_intervals}).

%% file: 04-evaluation.tex
\section{Benchmark Results}
\label{sec:evaluation}

We organize the results around the four questions in Section~\ref{sec:introduction}. Section~\ref{sec:target_tuning} measures the seven-dimensional change after Camera-view and Sensor-noise tuning of all five models, and Section~\ref{sec:source_domain_control} tests whether the resulting regressions reflect forgetting of the unperturbed tasks. Section~\ref{sec:data_coverage} then broadens the training data, and Section~\ref{sec:mitigation} compares tuning scope, weight interpolation, and feature preservation on X-VLA and OpenVLA-OFT. Each subsection ends with its finding.


\subsection{Does Targeted Vision Tuning Affect Off-Target Robustness?}
\label{sec:target_tuning}

Table~\ref{tab:targeted_tuning_results} reports absolute target and overall success rates before and after tuning together with the four gains in Equation~\ref{eq:retention_metrics}. Targeted tuning generally improves robustness to the selected perturbation, so we focus on how much performance each model preserves on the remaining dimensions.

\begin{table}[t]
  \caption{Target, off-target, and overall results under targeted vision tuning. Success rates are reported as percentages and signed gains are reported in percentage points.}
  \label{tab:targeted_tuning_results}
  \centering
  \setlength{\tabcolsep}{2.4pt}
  \renewcommand{\arraystretch}{1.03}
  \begin{tabular*}{\linewidth}{@{\extracolsep{\fill}}lrrrrrrrr@{}}
    \toprule
    & \multicolumn{3}{c}{\textbf{Target success}} & \multicolumn{2}{c}{\textbf{Off-target change}} & \multicolumn{3}{c}{\textbf{Overall success}} \\
    \cmidrule(lr){2-4}\cmidrule(lr){5-6}\cmidrule(lr){7-9}
    \textbf{Model} & \textbf{Base} & \textbf{Tuned} & $\boldsymbol{G^{\mathrm{target}}}$ & $\boldsymbol{G^{\mathrm{off}}}$ & $\boldsymbol{G^{\mathrm{worst}}}$ & \textbf{Base} & \textbf{Tuned} & $\boldsymbol{G^{\mathrm{all}}}$ \\
    \midrule
    \multicolumn{9}{c}{\textbf{Camera-view tuning}} \\
    \hline
    OpenVLA-OFT & $55.2$ & $94.5$ & $+39.3$ & $+1.4$ & $-2.8$ & $68.3$ & $75.8$ & $+7.5$ \\
    $\pi_{0.5}$ & $70.7$ & $92.7$ & $+22.0$ & $-0.3$ & $-3.9$ & $84.0$ & $87.0$ & $+3.0$ \\
    InternVLA-A1.5 & $80.8$ & $83.4$ & $+2.6$ & $-5.2$ & $-12.4$ & $83.7$ & $79.5$ & $-4.2$ \\
    NORA-1.5 & $6.0$ & $34.7$ & $+28.7$ & $-10.1$ & $-17.8$ & $47.4$ & $43.8$ & $-3.7$ \\
    X-VLA & $19.6$ & $89.0$ & $+69.4$ & $-4.6$ & $-10.7$ & $69.3$ & $76.3$ & $+7.0$ \\
    \midrule
    \multicolumn{9}{c}{\textbf{Sensor-noise tuning}} \\
    \hline
    OpenVLA-OFT & $72.8$ & $92.6$ & $+19.8$ & $+0.3$ & $-0.6$ & $68.3$ & $71.7$ & $+3.4$ \\
    $\pi_{0.5}$ & $86.5$ & $96.7$ & $+10.2$ & $-1.3$ & $-4.1$ & $84.0$ & $84.3$ & $+0.3$ \\
    InternVLA-A1.5 & $93.7$ & $91.8$ & $-1.9$ & $-2.5$ & $-6.0$ & $83.7$ & $81.2$ & $-2.6$ \\
    NORA-1.5 & $16.0$ & $61.3$ & $+45.3$ & $-1.0$ & $-4.4$ & $47.4$ & $53.6$ & $+6.2$ \\
    X-VLA & $60.9$ & $93.7$ & $+32.8$ & $-2.2$ & $-4.2$ & $69.3$ & $72.5$ & $+3.2$ \\
    \bottomrule
  \end{tabular*}
  \vspace{-0.4em}
\end{table}


\textbf{Camera-view tuning.} Camera-view tuning produces the largest contrast. X-VLA gains $69.4$ points on the target dimension but has $G^{\mathrm{off}}=-4.6$ and $G^{\mathrm{worst}}=-10.7$, while NORA-1.5 gains $28.7$ points on target but falls by $10.1$ points on average off target and by $17.8$ points on its worst dimension. In contrast, OpenVLA-OFT combines a $+39.3$ target gain with $G^{\mathrm{off}}=+1.4$ and $G^{\mathrm{all}}=+7.5$.

\textbf{Sensor-noise tuning.} Off-target regression also appears when models are tuned on Sensor-noise data, confirming that the effect is not unique to Camera-view tuning. As under Camera-view tuning, four of the five models have negative mean off-target gains, ranging from $-1.0$ to $-2.5$ points. OpenVLA-OFT remains the exception, retaining a positive mean off-target gain of $+0.3$ while improving Sensor-noise success by $19.8$ points.

\begin{figure}[t]
  \centering
  \includegraphics[width=0.95\linewidth]{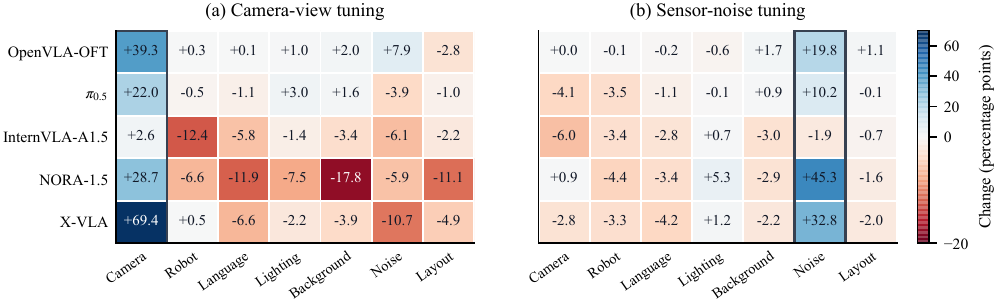}
  \vspace{-0.4em}
  \caption{\textbf{Cross-dimensional success rate changes after targeted vision tuning.} Each cell reports a percentage-point change. Blue denotes improvement, red denotes regression, and the outlined column marks the target perturbation.}
  \label{fig:cross_dimension_changes}
\end{figure}

\textbf{Cross-dimension analysis.} Figure~\ref{fig:cross_dimension_changes} expands these summaries into the complete seven-dimensional change vector. Language and object-layout success decline in nine of the ten model--domain configurations, by $3.7$ and $2.5$ points on average, respectively; robot-state success rates decline in eight configurations, by $3.3$ points on average. Cross-effects also occur between the two visual target domains: Sensor-noise success declines for four of five models after Camera-view tuning, while Camera-view success declines for three of five models after Sensor-noise tuning. Vision tuning can therefore reduce robustness outside the target domain, even to non-visual perturbations.

\textbf{Model dependence.} The magnitude and ordering of off-target regression change across both models and target domains. NORA-1.5 has $G^{\mathrm{off}}=-10.1$ after Camera-view tuning but $-1.0$ after Sensor-noise tuning, whereas OpenVLA-OFT retains positive mean off-target gains in both settings. Visual-module parameter share also has a Spearman rank correlation of $\rho=-0.70$ with mean off-target gain averaged over the two settings, relating a larger visual-module share to weaker retention.

\begin{figure}[t]
  \centering
  \includegraphics[width=\linewidth]{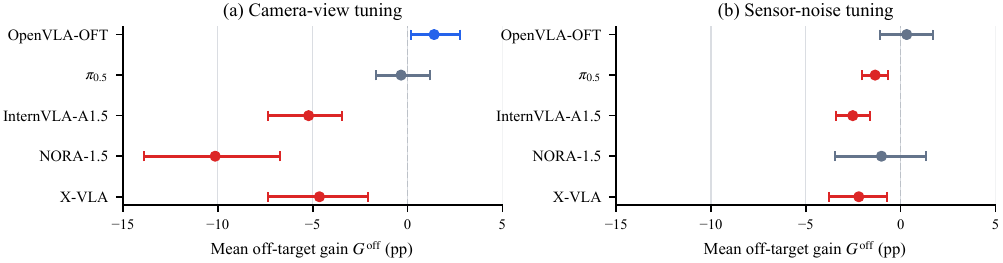}
  \vspace{-0.9em}
  \caption{\textbf{Statistical uncertainty in off-target retention.} Points show mean off-target gains, and horizontal lines show pointwise $95\%$ confidence intervals from a paired cluster bootstrap with 20,000 resamples over the 40 source tasks. Red intervals lie below zero, blue intervals lie above zero, and gray intervals overlap zero.}
  \label{fig:off_target_ci}
  \vspace{-1em}
\end{figure}

\textbf{Statistical uncertainty.} Pointwise $95\%$ confidence intervals across the 40 source tasks (Figure~\ref{fig:off_target_ci}) place $G^{\mathrm{off}}$ below zero in six of the ten model--domain settings and above zero only for OpenVLA-OFT under Camera-view tuning. Under Camera-view tuning, the intervals are $[-13.9,-6.7]$ for NORA-1.5, $[-7.4,-2.1]$ for X-VLA, and $[+0.2,+2.8]$ for OpenVLA-OFT. Overall-gain intervals lie below zero for InternVLA-A1.5 under both target domains and NORA-1.5 under Camera-view tuning, so off-target regressions can outweigh target gains.

\textbf{Why aggregate success is insufficient.} These results illustrate the first design principle in Section~\ref{sec:design_principles}: X-VLA's $+69.4$ target gain yields a positive overall gain of $+7.0$ despite regressions on five of six off-target dimensions, while OpenVLA-OFT's positive mean off-target gain still comes with a $-2.8$ drop on its worst dimension.

\begin{finding}{1}{Targeted vision tuning is not local}
Mean off-target gain is negative in 8 of 10 model--domain configurations. The target gain does not predict retention: under Camera-view tuning, X-VLA gains $69.4$ points on target but loses $4.6$ off target, whereas OpenVLA-OFT gains on both.
\end{finding}

\subsection{Are Off-Target Regressions Explained by Source-Domain Forgetting?}
\label{sec:source_domain_control}

{Off-target regressions could reflect a general loss on the original task distribution rather than reduced robustness to specific perturbations. We therefore evaluate each base--tuned pair on original LIBERO and compute the source-adjusted change $R_{m,c,d}=\Delta_{m,c,d}-(S^{\mathrm{tuned,orig}}_{m,c}-S^{\mathrm{base,orig}}_{m})$. A negative value indicates a larger decline under perturbation $d$ than on original LIBERO.}

\begin{figure}[t]
  \centering
  \includegraphics[width=0.95\linewidth]{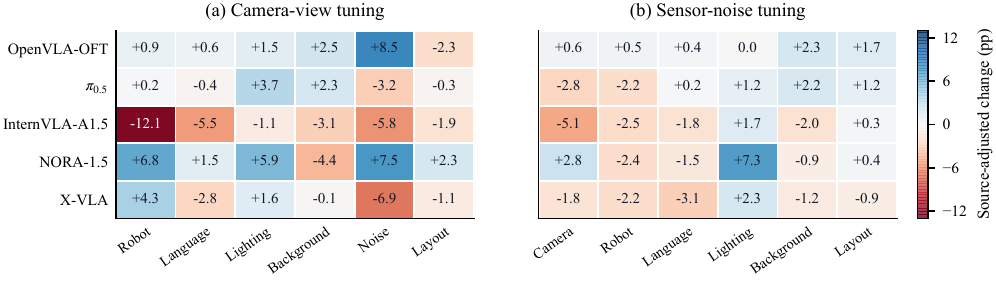}
  \vspace{-0.4em}
  \caption{{\textbf{Source-adjusted off-target changes.} Results for (a) Camera-view and (b) Sensor-noise tuning. Each cell subtracts the corresponding change on original LIBERO; negative values indicate additional perturbation-specific regression.}}
  \label{fig:source_domain_control}
\end{figure}

{Original LIBERO changes remain within $3.8$ percentage points for all checkpoints except NORA-1.5 after Camera-view tuning ($-13.4$ points). Figure~\ref{fig:source_domain_control} shows that every Camera-view-tuned model still regresses on at least one off-target perturbation after adjustment. The same holds for four models after Sensor-noise tuning, while OpenVLA-OFT remains effectively unchanged or improves. Appendix Table~\ref{tab:original_libero_retention} reports the absolute success rates.}

\begin{finding}{2}{Regressions are not explained only by source-domain forgetting}
After subtracting the change on original LIBERO, every Camera-view-tuned model and four of five Sensor-noise-tuned models still regress on at least one off-target perturbation.
\end{finding}

\vspace{-1em}
\subsection{Does Broader Training Coverage Prevent Off-Target Regression?}
\label{sec:data_coverage}

\begin{figure}[t]
  \centering
  \includegraphics[width=0.93\linewidth]{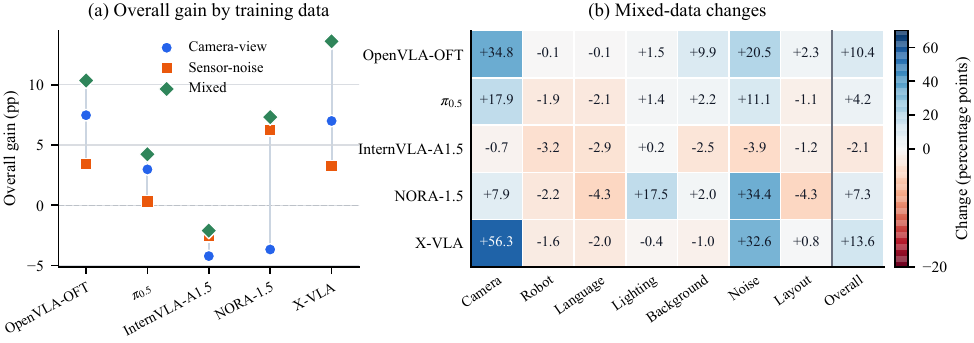}
  \vspace{-0.4em}
  \caption{\textbf{Effect of training-data coverage.} (a) Overall gains under Camera-view, Sensor-noise, and Mixed training. (b) Dimension-level and overall changes after Mixed training. }
  
  \label{fig:training_data_coverage}
  \vspace{-0.6em}
\end{figure}

{Target-specific data may promote specialization at the expense of retention. We therefore apply the same Dual-LoRA configuration and optimization budget to the complete \libplus{} training set, which covers every evaluation dimension except robot state and object layout over the same 40 manipulation tasks. As Mixed training has no single target dimension, Figure~\ref{fig:training_data_coverage} compares its overall gain with Camera-view and Sensor-noise training and reports its seven dimension-level changes.}

{Mixed training yields a higher overall gain than either target-specific setting for all five models, from $-2.1$ points for InternVLA-A1.5 to $+13.6$ for X-VLA. Yet every model still regresses on at least one dimension, and Camera-view gains are smaller than under Camera-view-specific training.}

\begin{finding}{3}{Broader coverage raises overall gain but does not remove regressions}
Mixed training yields a higher overall gain than either target-specific setting for all five models, yet every model still regresses on at least one dimension.
\end{finding}

\subsection{Can Alternative Tuning Strategies Control Cross-Dimension Effects?}
\label{sec:mitigation}

\begin{figure}[t]
  \centering
  \includegraphics[width=0.9\linewidth]{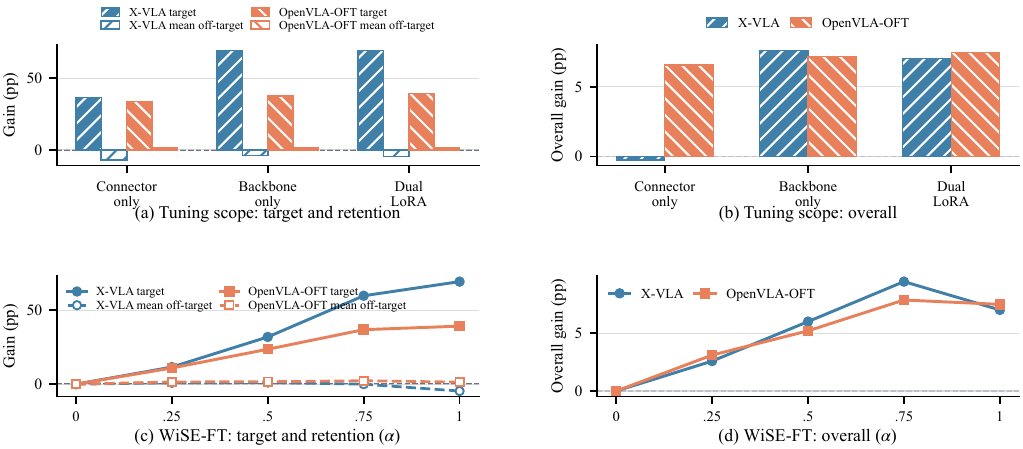}
  \vspace{-0.4em}
  \caption{\textbf{Target--retention trade-offs of tuning strategies.} (a--b) Tuning scope. (c--d) WiSE-FT.}
  \label{fig:tuning_strategy_results}
  \vspace{-1em}
\end{figure}

\textbf{Tuning scope.} Figure~\ref{fig:tuning_strategy_results}(a)--(b) compare Connector-only, Backbone-only, and Dual LoRA on X-VLA and OpenVLA-OFT. For X-VLA, Backbone-only matches the $+69.4$ target gain of Dual LoRA while improving mean off-target gain from $-4.6$ to $-3.9$ and overall gain from $+7.0$ to $+7.6$. OpenVLA-OFT instead performs best with Dual LoRA, reaching a $+39.3$ target gain, $+1.4$ mean off-target gain, and $+7.5$ overall gain. The preferred tuning scope therefore differs across models.

\textbf{Weight interpolation.} Figure~\ref{fig:tuning_strategy_results}(c)--(d) show how WiSE-FT changes the balance between target adaptation and off-target retention. At $\alpha=0.75$, X-VLA's target gain decreases from $+69.4$ to $+59.8$, while its mean off-target gain improves from $-4.6$ to $0.0$ and its overall gain increases from $+7.0$ to $+9.4$. OpenVLA-OFT shows the same trade-off, with target gain decreasing from $+39.3$ to $+36.9$ and mean off-target gain increasing from $+1.4$ to $+2.3$. Weight interpolation therefore improves retention and overall performance at the cost of a smaller target-domain gain.

\begin{table}[t]
  \caption{Camera-view results for Dual LoRA and DELTA-style feature preservation.}
  \label{tab:delta_results}
  \centering
  \setlength{\tabcolsep}{8pt}
  \renewcommand{\arraystretch}{1.02}
  \begin{tabular*}{\linewidth}{@{\extracolsep{\fill}}llrrr@{}}
    \toprule
    \textbf{Model} & \textbf{Strategy} & $\boldsymbol{G^{\mathrm{target}}}$ & $\boldsymbol{G^{\mathrm{off}}}$ & $\boldsymbol{G^{\mathrm{all}}}$ \\
    \midrule
    X-VLA & Dual LoRA & $+69.4$ & $-4.6$ & $+7.0$ \\
    X-VLA & DELTA-style & $+62.7$ & $+1.2$ & $+10.9$ \\ \hline
    OpenVLA-OFT & Dual LoRA & $+39.3$ & $+1.4$ & $+7.5$ \\
    OpenVLA-OFT & DELTA-style & $+13.6$ & $-23.1$ & $-16.6$ \\
    \bottomrule
  \end{tabular*}
\end{table}

\textbf{Feature preservation.} Table~\ref{tab:delta_results} compares DELTA-style with the reference Dual-LoRA configuration. DELTA-style retains a $+62.7$ Camera-view gain on X-VLA, is positive on five of six off-target dimensions, and raises overall gain from $+7.0$ to $+10.9$. On OpenVLA-OFT it regresses on all six off-target dimensions, producing $G^{\mathrm{off}}=-23.1$ and $G^{\mathrm{all}}=-16.6$.

\begin{finding}{4}{Tuning strategies are model-dependent}
No strategy is best for both models: DELTA-style feature preservation raises X-VLA's mean off-target gain from $-4.6$ to $+1.2$ points but lowers OpenVLA-OFT's from $+1.4$ to $-23.1$.
\end{finding}

%% file: 05-conclusion.tex
\section{Conclusion} 


We introduced \bench{}, a paired benchmark for measuring how targeted vision tuning changes a VLA's robustness across perturbation dimensions. Across five open VLAs, targeted tuning generally improves the target perturbation, but its effects are not local: off-target regressions are common, model-dependent, not predicted by the target gain, and not explained only by source-domain forgetting. Broader training coverage and alternative tuning strategies move this trade-off without removing it. VLA adaptation studies should report target improvement and off-target retention jointly.


\section*{Reproducibility statement}
We release a public repository at \url{https://github.com/BoruiXu/VLA_vision_tuning_benchmark} that contains the training, merging, serving, and evaluation code for all five models, the per-episode outcomes of every reported run, and scripts that recompute the reported metrics, confidence intervals, and result figures. And all model base checkpoints and the \libplus{} data are publicly available.

%% file: appendix.tex
\clearpage
\section{Additional Benchmark Details}
\label{app:benchmark_details}

\subsection{Tuning Configurations}

Table~\ref{tab:tuning_strategies} summarizes the visual tuning configurations compared in Section~\ref{sec:mitigation}.

\begin{table}[h]
  \caption{Visual tuning configurations used in the strategy comparison.}
  \label{tab:tuning_strategies}
  \centering
  \setlength{\tabcolsep}{4pt}
  \renewcommand{\arraystretch}{1.04}
  \begin{tabularx}{\linewidth}{@{}p{0.16\linewidth}p{0.25\linewidth}X@{}}
    \hline
    \textbf{Category} & \textbf{Configuration} & \textbf{Operation} \\
    \hline
    Reference & Dual LoRA & Rank-32 LoRA on visual backbone and connector \\ \hline
    Tuning scope & Backbone-only LoRA & Rank-32 LoRA on visual backbone; connector frozen \\
    Tuning scope & Connector-only LoRA & Rank-32 LoRA on connector; visual backbone frozen \\ \hline
    Retention & WiSE-FT & Base--Dual-LoRA visual-weight interpolation \\
    Retention & DELTA-style & Dual LoRA with post-connector feature preservation \\
    \bottomrule
  \end{tabularx}
\end{table}

\subsection{Model Parameter Scales}

We obtain parameter counts from the released checkpoints after identifying the visual backbone and connector defined in Section~\ref{sec:benchmark}. The visual count describes how much of each architecture lies on the visual pathway; it is distinct from the number of LoRA parameters optimized during fine-tuning.

\begin{figure}[h]
  \centering
  \includegraphics[width=\linewidth]{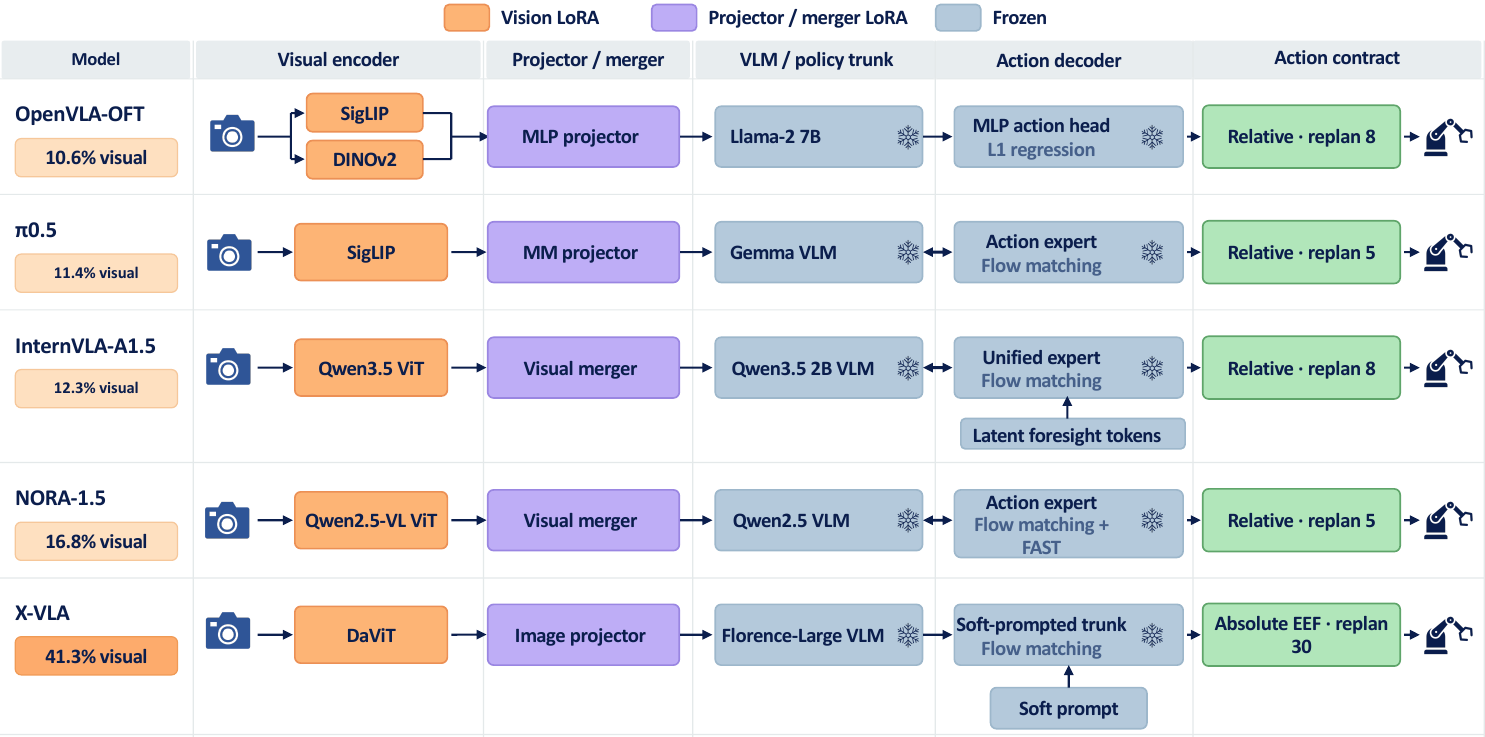}
  \caption{\textbf{Architectures and visual interfaces of the five VLA models.} The reference configuration updates the visual backbone and connector and freezes the multimodal backbone and action module.}
  \label{fig:architecture_comparison}
\end{figure}

\begin{table}[h]
  \caption{Model parameter scales. Vision parameters include the visual backbone and connector.}
  \label{tab:model_parameters}
  \centering
  \small
  \setlength{\tabcolsep}{10pt}
  \renewcommand{\arraystretch}{1.05}
  \begin{tabular}{lrrr}
    \toprule
    \textbf{Model} & \textbf{Vision params.} & \textbf{Total params.} & \textbf{Vision share} \\
    \midrule
    OpenVLA-OFT & 802.3M & 7.54B & 10.64\% \\
    $\pi_{0.5}$ & 412.4M & 3.62B & 11.40\% \\
    InternVLA-A1.5 & 331.4M & 2.69B & 12.34\% \\
    NORA-1.5 & 668.7M & 3.99B & 16.77\% \\
    X-VLA & 362.8M & 0.88B & 41.26\% \\
    \bottomrule
  \end{tabular}
\end{table}

\subsection{Training Hyperparameters}

The reference runs share the training budget and adapter configuration described in Section~\ref{sec:benchmark}: BF16 precision, effective batch size 32, 12,000 optimizer updates, 384,000 sampled examples, rank-32 LoRA with $\alpha=16$ and zero dropout, and checkpoints every 2,000 updates. Training-based controlled comparisons retain the corresponding model's budget, optimizer, and schedule while changing only the designated data or visual tuning scope; WiSE-FT operates on completed checkpoints and requires no additional optimization. We retain the optimizer and schedule values from each model's official repository instead of imposing one learning rate across architectures. Table~\ref{tab:training_hyperparameters} separates optimizer settings from the learning-rate schedule and execution contract.

\begin{table}[h]
  \caption{Model-specific optimization, schedule, and execution settings.}
  \label{tab:training_hyperparameters}
  \centering
  \small
  \textbf{Optimizer settings}\\[3pt]
  \begin{tabular}{lcccc}
    \toprule
    \textbf{Model} & \textbf{Learning rate} & \textbf{$(\beta_1,\beta_2)$} & \textbf{Weight decay} & \textbf{Grad. clip} \\
    \midrule
    OpenVLA-OFT & $5\!\times\!10^{-4}$ & $(0.9,0.999)$ & $0.1$ & None \\
    $\pi_{0.5}$ & $5\!\times\!10^{-5}$ & $(0.9,0.95)$ & $10^{-10}$ & 1.0 \\
    InternVLA-A1.5 & $5\!\times\!10^{-5}$ & $(0.9,0.95)$ & $0.01$ & 1.0 \\
    NORA-1.5 & $8\!\times\!10^{-5}$ & $(0.9,0.95)$ & $10^{-8}$ & 1.0 \\
    X-VLA & $10^{-4}$ & $(0.9,0.95)$ & 0 & 1.0 \\
    \bottomrule
  \end{tabular}
  \vspace{8pt}

  \textbf{Schedule and execution settings}\\[3pt]
  \begin{tabularx}{0.84\linewidth}{@{}lc>{\raggedright\arraybackslash}Xc@{}}
    \toprule
    \textbf{Model} & \textbf{Warmup} & \textbf{Post-warmup schedule} & \textbf{Horizon} \\
    \midrule
    OpenVLA-OFT & 1k & Cosine decay to 0 & 8 \\
    $\pi_{0.5}$ & 1k & Constant & 5 \\
    InternVLA-A1.5 & 2k & Cosine decay to $5\!\times\!10^{-6}$ & 8 \\
    NORA-1.5 & 1k & Cosine decay to 0 & 5 \\
    X-VLA & 2k & Cosine decay to $10^{-5}$ & 30 \\
    \bottomrule
  \end{tabularx}
\end{table}

The execution horizon is the number of actions applied before the VLA model is queried again. All other observation preprocessing, normalization, action-space, and gripper settings follow the model-specific contracts used by \texttt{vla-evaluation-harness}~\citep{choi2026vlaeval}.

\subsection{{Original LIBERO Retention}}

{Table~\ref{tab:original_libero_retention} reports the source-domain results used in Section~\ref{sec:source_domain_control}. Each checkpoint is evaluated on the same 2,000 original LIBERO episodes, covering 40 tasks with 50 fixed initial states per task.}

\begin{table}[]
  \caption{{Original LIBERO success before and after targeted vision tuning. Success rates are percentages and changes are percentage points.}}
  \label{tab:original_libero_retention}
  \centering
  \setlength{\tabcolsep}{5pt}
  \renewcommand{\arraystretch}{1.04}
  {
  \begin{tabular*}{\linewidth}{@{\extracolsep{\fill}}lrrrrr@{}}
    \toprule
    & \textbf{Base} & \multicolumn{2}{c}{\textbf{Camera-view tuned}} & \multicolumn{2}{c}{\textbf{Sensor-noise tuned}} \\
    \cmidrule(lr){3-4}\cmidrule(lr){5-6}
    \textbf{Model} & \textbf{Success (\%)} & \textbf{Success (\%)} & \textbf{Change (pp)} & \textbf{Success (\%)} & \textbf{Change (pp)} \\
    \midrule
    OpenVLA-OFT & $97.20$ & $96.65$ & $-0.55$ & $96.60$ & $-0.60$ \\
    $\pi_{0.5}$ & $97.35$ & $96.65$ & $-0.70$ & $96.05$ & $-1.30$ \\
    InternVLA-A1.5 & $98.25$ & $97.95$ & $-0.30$ & $97.30$ & $-0.95$ \\
    NORA-1.5 & $94.70$ & $81.30$ & $-13.40$ & $92.75$ & $-1.95$ \\
    X-VLA & $97.50$ & $93.70$ & $-3.80$ & $96.45$ & $-1.05$ \\
    \bottomrule
  \end{tabular*}
  }
\end{table}

\subsection{DELTA-Style Feature Preservation}
\label{app:delta_style}

DELTA-style adapts DELTA's behavior-regularization principle~\citep{li2018delta} to VLA vision tuning. For every Camera-view training batch, the current model acts as a student and an untouched copy of the base visual module acts as a frozen teacher. Both branches receive the same augmented image, the student follows the standard action-training path, and the teacher produces only a reference visual representation. The student updates the same rank-32 LoRA parameters on the visual backbone and connector as the reference configuration; the teacher remains outside the optimizer and checkpoint. We replay the student's random-number-generator state for the teacher forward so that stochastic visual operations use the same masks in both branches.

Table~\ref{tab:delta_implementation} shows the model-dependent feature taps. X-VLA matches the output of \texttt{vlm.image\_proj\_norm}, after its image projector and normalization, against the corresponding output of a frozen copy of the base VLM visual path. OpenVLA-OFT matches the projector output against a frozen sequential copy of its base vision backbone and projector. The language backbone and action module are not duplicated and remain frozen in the student model.

\begin{table}[h]
  \caption{Model-specific feature taps for DELTA-style preservation.}
  \label{tab:delta_implementation}
  \centering
  \setlength{\tabcolsep}{7pt}
  \renewcommand{\arraystretch}{1.04}
  \begin{tabularx}{\linewidth}{@{}l>{\raggedright\arraybackslash}X>{\raggedright\arraybackslash}p{0.34\linewidth}@{}}
    \toprule
    \textbf{Model} & \textbf{Frozen teacher} & \textbf{Matched representation} \\
    \midrule
    X-VLA & Base visual path and image projector & \texttt{image\_proj\_norm} output \\
    OpenVLA-OFT & Base vision backbone and projector & Projector output \\
    \bottomrule
  \end{tabularx}
\end{table}

Let $H,H^0\in\mathbb{R}^{B\times N\times C}$ denote the student and teacher post-connector representations, where $B$ is the number of valid images, $N$ the number of visual tokens, and $C$ the feature dimension. For channel $c$, we compute its mean squared change and first-order sensitivity to the action loss:
\begin{equation}
e_c=\frac{1}{BN}\sum_{b=1}^{B}\sum_{n=1}^{N}\left(H_{b,n,c}-H^0_{b,n,c}\right)^2, \qquad q_c=\frac{1}{BN}\sum_{b=1}^{B}\sum_{n=1}^{N}\left|H_{b,n,c}\frac{\partial\mathcal{L}_{\mathrm{action}}}{\partial H_{b,n,c}}\right|.
\label{eq:delta_error_importance}
\end{equation}
The sensitivity $q_c$ is the first-order Taylor approximation to the change in action loss caused by suppressing channel $c$. We normalize these sensitivities by their mean and convert them into non-negative attention weights,
\begin{equation}
\bar q_c=\frac{q_c}{\max\!\left(C^{-1}\sum_{k=1}^{C}q_k,\epsilon\right)}, \qquad a_c=C\frac{\exp(\bar q_c/\tau)}{\sum_{k=1}^{C}\exp(\bar q_k/\tau)}, \qquad \epsilon=10^{-12},
\label{eq:delta_attention}
\end{equation}
so that the mean channel weight is one. The attention weights are detached before the preservation loss is differentiated, avoiding second-order gradients. The final objectives are
\begin{equation}
\mathcal{L}_{\mathrm{feature}}=\frac{1}{C}\sum_{c=1}^{C}a_c e_c, \qquad \mathcal{L}_{\mathrm{total}}=\mathcal{L}_{\mathrm{action}}+\lambda\mathcal{L}_{\mathrm{feature}},
\label{eq:delta_objective}
\end{equation}
with $\lambda=1$ and temperature $\tau=1$ in both configurations. This objective gives stronger preservation pressure to post-connector channels on which the current action loss is more sensitive, while letting less influential channels change more freely. After training, the teacher branch is discarded and the merged model matches the inference structure and cost of the reference Dual-LoRA model.

We call this configuration DELTA-style rather than DELTA because it changes three elements of the original algorithm. DELTA obtains supervised attention by explicitly ablating each source-network CNN filter and measuring the resulting target-task loss increase; our implementation replaces those repeated forwards with the Taylor approximation in Equation~\ref{eq:delta_error_importance}. DELTA regularizes convolutional feature maps and includes an additional penalty for parameters private to the target network, whereas our implementation matches one post-connector visual representation and adds no private-parameter penalty. Finally, the channel weights here are computed from the current student's VLA action loss and averaged across the batch and visual-token dimensions, rather than computed per example from exact source-filter ablations. The results in Table~\ref{tab:delta_results} therefore characterize this VLA-specific feature-anchoring configuration, not an exact reproduction of the original DELTA algorithm.

\input{appendix_detailed_results}

\input{appendix_strategy_results}

\input{appendix_ci}

\section{Qualitative Paired Rollouts}
\label{app:qualitative_cases}

\begin{figure}[h]
  \centering
  \includegraphics[width=\linewidth]{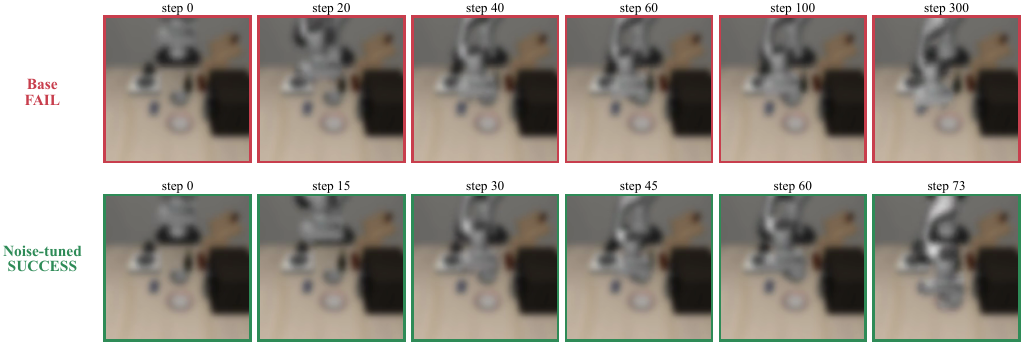}
  \caption{\textbf{Paired X-VLA trajectories under a Sensor-noise perturbation.} Six ordered frames are shown for the base failure (top) and {Sensor-noise-tuned} success (bottom) on the same episode. Frame labels give simulator steps.}
  \label{fig:qualitative_cases}
\end{figure}

Figure~\ref{fig:qualitative_cases} visualizes a paired X-VLA episode under Sensor-noise evaluation with the instruction {\texttt{put the bowl on the plate}}. The base and {Sensor-noise-tuned} models receive the same task, initial state, and noise perturbation. We show six temporally ordered frames from each trajectory, with one rollout per row. {The base model moves toward the bowl but does not place it on the plate before the 300-step limit. The tuned model grasps the bowl, moves it onto the plate, and succeeds after 73 steps.} The paired rows illustrate how an aggregate outcome reversal arises from different action sequences within the same evaluation condition.

\begin{figure}[h]
  \centering
  \includegraphics[width=\linewidth]{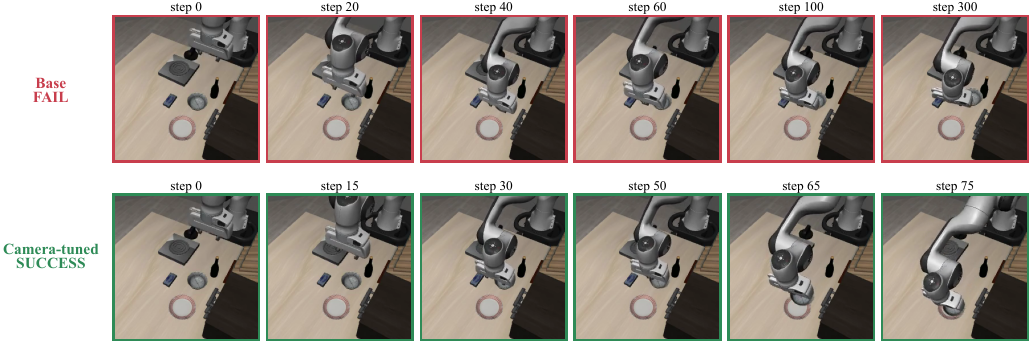}
  \caption{\textbf{Paired X-VLA trajectories under a Camera-view perturbation.} Six ordered frames are shown for the base failure (top) and Camera-view-tuned success (bottom) on the same episode. Frame labels give simulator steps.}
  \label{fig:qualitative_camera_view}
\end{figure}

Figure~\ref{fig:qualitative_camera_view} provides a second paired X-VLA example {for the same instruction} under a Camera-view perturbation. Both models receive the same changed viewpoint, initial state, and evaluation seed. The base model approaches the bowl but leaves it off the plate when it reaches the 300-step limit. The Camera-view-tuned model completes the transfer and succeeds after 75 steps. Six ordered frames from each trajectory show how target-domain tuning changes the execution of the same episode.

%% file: appendix_detailed_results.tex
\section{Fine-Grained Results by Task Suite}
Tables~\ref{tab:camera_suite_results} and~\ref{tab:noise_suite_results} expand the core reference matrix into the four LIBERO task suites whose evaluation composition is reported in Table~\ref{tab:evaluation_set}. Each cell reports the success rate of the original and tuned model on exactly the same episode keys, with the target perturbation shown in bold. The final column pools the four suites within the indicated perturbation and therefore matches the dimension-level rates underlying Figure~\ref{fig:cross_dimension_changes}.

\begin{table}[h]
  \caption{Camera-view tuning results by task suite. Each cell is Base/Tuned (\%).}
  \label{tab:camera_suite_results}
  \centering
\setlength{\tabcolsep}{4pt}
\renewcommand{\arraystretch}{0.90}
\begin{tabularx}{\linewidth}{@{}l*{5}{>{\centering\arraybackslash}X}@{}}
\toprule
\textbf{Perturbation} & \textbf{Spatial} & \textbf{Object} & \textbf{Goal} & \textbf{LIBERO-10} & \textbf{All suites} \\
\midrule
\multicolumn{6}{c}{\textbf{OpenVLA-OFT}} \\
\hline
\textbf{Camera view} & $48.7/98.9$ & $67.9/99.7$ & $59.3/93.6$ & $45.1/86.4$ & $55.2/94.5$ \\
Robot state & $23.1/24.6$ & $19.8/19.1$ & $21.3/21.8$ & $36.6/36.9$ & $25.2/25.5$ \\
Language & $95.9/96.4$ & $99.7/98.6$ & $58.0/59.0$ & $95.3/95.0$ & $86.5/86.6$ \\
Lighting & $97.9/98.6$ & $99.3/100.0$ & $97.1/93.9$ & $85.0/90.9$ & $95.0/96.0$ \\
Background & $88.0/88.4$ & $87.5/81.0$ & $84.0/88.6$ & $83.0/91.0$ & $85.5/87.5$ \\
Sensor noise & $75.5/80.6$ & $91.0/94.1$ & $64.6/76.3$ & $60.4/71.9$ & $72.8/80.7$ \\
Object layout & $88.3/82.6$ & $76.2/74.2$ & $50.4/49.9$ & $69.9/66.3$ & $70.8/67.9$ \\
\hline
\addlinespace[2pt]
\multicolumn{6}{c}{\textbf{$\pi_{0.5}$}} \\
\hline
\textbf{Camera view} & $76.6/92.8$ & $84.1/98.0$ & $74.5/93.6$ & $48.9/86.6$ & $70.7/92.7$ \\
Robot state & $83.1/79.4$ & $70.9/72.1$ & $72.6/73.6$ & $76.3/75.3$ & $75.5/75.0$ \\
Language & $92.1/88.2$ & $91.8/91.5$ & $70.7/66.3$ & $89.6/94.0$ & $85.7/84.6$ \\
Lighting & $95.5/100.0$ & $97.6/98.3$ & $93.9/97.5$ & $94.2/97.4$ & $95.4/98.3$ \\
Background & $97.7/97.7$ & $98.4/98.4$ & $90.7/95.4$ & $91.7/93.1$ & $94.4/96.0$ \\
Sensor noise & $92.3/87.7$ & $93.6/91.0$ & $88.1/84.2$ & $73.9/69.3$ & $86.5/82.6$ \\
Object layout & $98.4/95.6$ & $90.6/90.6$ & $72.0/71.8$ & $86.9/85.6$ & $86.6/85.6$ \\
\hline
\addlinespace[2pt]
\multicolumn{6}{c}{\textbf{InternVLA-A1.5}} \\
\hline
\textbf{Camera view} & $89.9/88.0$ & $98.2/96.7$ & $68.6/66.9$ & $68.0/82.8$ & $80.8/83.4$ \\
Robot state & $57.7/49.1$ & $56.5/38.7$ & $48.4/39.4$ & $50.1/36.4$ & $53.0/40.6$ \\
Language & $85.1/80.0$ & $93.5/92.9$ & $74.9/68.8$ & $95.8/84.9$ & $87.0/81.2$ \\
Lighting & $90.1/90.8$ & $99.7/97.3$ & $97.1/97.5$ & $98.9/94.5$ & $96.4/95.0$ \\
Background & $98.8/96.9$ & $100.0/100.0$ & $97.9/92.9$ & $97.9/91.7$ & $98.6/95.2$ \\
Sensor noise & $91.7/90.6$ & $99.8/99.1$ & $85.0/71.5$ & $96.9/88.0$ & $93.7/87.6$ \\
Object layout & $96.6/90.6$ & $85.1/83.4$ & $69.2/68.0$ & $88.8/89.4$ & $84.3/82.2$ \\
\hline
\addlinespace[2pt]
\multicolumn{6}{c}{\textbf{NORA-1.5}} \\
\hline
\textbf{Camera view} & $9.0/42.0$ & $3.5/44.2$ & $6.4/31.9$ & $5.3/22.0$ & $6.0/34.7$ \\
Robot state & $54.0/50.9$ & $49.7/39.9$ & $33.7/36.4$ & $43.0/26.7$ & $44.8/38.1$ \\
Language & $78.7/72.3$ & $79.4/74.6$ & $49.0/43.9$ & $74.9/43.6$ & $70.0/58.1$ \\
Lighting & $90.8/79.1$ & $26.6/40.7$ & $82.1/62.7$ & $46.7/32.1$ & $61.4/53.9$ \\
Background & $79.8/69.4$ & $79.8/74.6$ & $89.0/60.1$ & $55.7/31.5$ & $75.7/58.0$ \\
Sensor noise & $13.1/7.7$ & $23.9/18.2$ & $14.0/8.4$ & $12.5/5.8$ & $16.0/10.1$ \\
Object layout & $91.4/71.2$ & $67.7/68.7$ & $65.2/52.5$ & $69.6/56.1$ & $73.4/62.2$ \\
\hline
\addlinespace[2pt]
\multicolumn{6}{c}{\textbf{X-VLA}} \\
\hline
\textbf{Camera view} & $18.4/95.5$ & $16.7/96.5$ & $23.5/89.0$ & $19.6/76.1$ & $19.6/89.0$ \\
Robot state & $91.7/89.7$ & $94.5/90.2$ & $88.5/89.2$ & $78.1/85.2$ & $88.1/88.6$ \\
Language & $81.8/71.0$ & $91.5/84.5$ & $55.6/48.5$ & $65.5/64.2$ & $73.0/66.4$ \\
Lighting & $100.0/97.9$ & $92.6/98.7$ & $98.6/90.0$ & $95.6/90.9$ & $96.7/94.5$ \\
Background & $100.0/91.9$ & $98.0/96.4$ & $91.5/91.1$ & $91.7/86.2$ & $95.1/91.2$ \\
Sensor noise & $59.0/45.6$ & $50.2/46.4$ & $66.5/62.3$ & $67.7/47.0$ & $60.9/50.2$ \\
Object layout & $91.2/82.9$ & $51.9/57.3$ & $62.6/58.1$ & $70.8/56.1$ & $68.7/63.7$ \\
\bottomrule
\end{tabularx}
\end{table}

\begin{table}[h]
  \caption{Sensor-noise tuning results by task suite. Each cell is Base/Tuned (\%).}
  \label{tab:noise_suite_results}
  \centering
\setlength{\tabcolsep}{4pt}
\renewcommand{\arraystretch}{0.90}
\begin{tabularx}{\linewidth}{@{}l*{5}{>{\centering\arraybackslash}X}@{}}
\toprule
\textbf{Perturbation} & \textbf{Spatial} & \textbf{Object} & \textbf{Goal} & \textbf{LIBERO-10} & \textbf{All suites} \\
\midrule
\multicolumn{6}{c}{\textbf{OpenVLA-OFT}} \\
\hline
Camera view & $48.7/53.2$ & $67.9/73.2$ & $59.3/55.6$ & $45.1/39.6$ & $55.2/55.2$ \\
Robot state & $23.1/20.0$ & $19.8/21.6$ & $21.3/21.0$ & $36.6/37.7$ & $25.2/25.2$ \\
Language & $95.9/93.8$ & $99.7/99.4$ & $58.0/59.3$ & $95.3/95.6$ & $86.5/86.3$ \\
Lighting & $97.9/96.9$ & $99.3/100.0$ & $97.1/89.2$ & $85.0/90.9$ & $95.0/94.4$ \\
Background & $88.0/89.1$ & $87.5/90.3$ & $84.0/84.3$ & $83.0/85.5$ & $85.5/87.2$ \\
\textbf{Sensor noise} & $75.5/92.3$ & $91.0/97.4$ & $64.6/90.8$ & $60.4/89.8$ & $72.8/92.6$ \\
Object layout & $88.3/91.2$ & $76.2/75.9$ & $50.4/50.4$ & $69.9/72.1$ & $70.8/71.9$ \\
\hline
\addlinespace[2pt]
\multicolumn{6}{c}{\textbf{$\pi_{0.5}$}} \\
\hline
Camera view & $76.6/70.2$ & $84.1/79.3$ & $74.5/71.8$ & $48.9/46.1$ & $70.7/66.5$ \\
Robot state & $83.1/82.9$ & $70.9/65.3$ & $72.6/69.4$ & $76.3/71.5$ & $75.5/71.9$ \\
Language & $92.1/91.0$ & $91.8/92.7$ & $70.7/67.6$ & $89.6/88.8$ & $85.7/84.6$ \\
Lighting & $95.5/96.2$ & $97.6/97.6$ & $93.9/97.1$ & $94.2/89.8$ & $95.4/95.3$ \\
Background & $97.7/97.7$ & $98.4/98.4$ & $90.7/93.6$ & $91.7/92.4$ & $94.4/95.4$ \\
\textbf{Sensor noise} & $92.3/97.2$ & $93.6/97.9$ & $88.1/98.4$ & $73.9/93.8$ & $86.5/96.7$ \\
Object layout & $98.4/97.9$ & $90.6/91.3$ & $72.0/70.8$ & $86.9/87.5$ & $86.6/86.5$ \\
\hline
\addlinespace[2pt]
\multicolumn{6}{c}{\textbf{InternVLA-A1.5}} \\
\hline
Camera view & $89.9/82.7$ & $98.2/88.6$ & $68.6/64.0$ & $68.0/65.2$ & $80.8/74.8$ \\
Robot state & $57.7/54.9$ & $56.5/51.5$ & $48.4/43.8$ & $50.1/49.1$ & $53.0/49.6$ \\
Language & $85.1/85.4$ & $93.5/91.8$ & $74.9/69.0$ & $95.8/92.2$ & $87.0/84.2$ \\
Lighting & $90.1/96.6$ & $99.7/98.7$ & $97.1/96.8$ & $98.9/96.4$ & $96.4/97.1$ \\
Background & $98.8/91.9$ & $100.0/98.8$ & $97.9/94.7$ & $97.9/97.2$ & $98.6/95.6$ \\
\textbf{Sensor noise} & $91.7/87.5$ & $99.8/98.8$ & $85.0/84.7$ & $96.9/94.4$ & $93.7/91.8$ \\
Object layout & $96.6/92.7$ & $85.1/84.6$ & $69.2/69.2$ & $88.8/91.0$ & $84.3/83.7$ \\
\hline
\addlinespace[2pt]
\multicolumn{6}{c}{\textbf{NORA-1.5}} \\
\hline
Camera view & $9.0/9.6$ & $3.5/5.6$ & $6.4/6.6$ & $5.3/6.0$ & $6.0/6.9$ \\
Robot state & $54.0/50.0$ & $49.7/48.7$ & $33.7/26.9$ & $43.0/37.4$ & $44.8/40.4$ \\
Language & $78.7/69.5$ & $79.4/79.4$ & $49.0/46.6$ & $74.9/73.1$ & $70.0/66.6$ \\
Lighting & $90.8/89.0$ & $26.6/44.8$ & $82.1/76.0$ & $46.7/57.3$ & $61.4/66.7$ \\
Background & $79.8/80.2$ & $79.8/78.2$ & $89.0/74.4$ & $55.7/60.2$ & $75.7/72.9$ \\
\textbf{Sensor noise} & $13.1/56.7$ & $23.9/78.2$ & $14.0/52.0$ & $12.5/56.8$ & $16.0/61.3$ \\
Object layout & $91.4/88.8$ & $67.7/66.0$ & $65.2/63.1$ & $69.6/70.2$ & $73.4/71.8$ \\
\hline
\addlinespace[2pt]
\multicolumn{6}{c}{\textbf{X-VLA}} \\
\hline
Camera view & $18.4/17.0$ & $16.7/11.1$ & $23.5/19.1$ & $19.6/19.6$ & $19.6/16.8$ \\
Robot state & $91.7/89.4$ & $94.5/91.0$ & $88.5/84.4$ & $78.1/75.1$ & $88.1/84.8$ \\
Language & $81.8/74.6$ & $91.5/88.4$ & $55.6/51.7$ & $65.5/63.2$ & $73.0/68.8$ \\
Lighting & $100.0/99.7$ & $92.6/100.0$ & $98.6/99.6$ & $95.6/92.0$ & $96.7/97.9$ \\
Background & $100.0/100.0$ & $98.0/98.4$ & $91.5/87.9$ & $91.7/86.5$ & $95.1/92.8$ \\
\textbf{Sensor noise} & $59.0/95.7$ & $50.2/95.7$ & $66.5/93.9$ & $67.7/90.0$ & $60.9/93.7$ \\
Object layout & $91.2/90.6$ & $51.9/48.9$ & $62.6/60.5$ & $70.8/68.6$ & $68.7/66.7$ \\
\bottomrule
\end{tabularx}
\end{table}

\clearpage

%% file: appendix_strategy_results.tex
\section{Dimension-Level Results for Tuning Strategies}
\label{app:strategy_dimension_results}
Section~\ref{sec:mitigation} reports the principal aggregate comparisons among tuning strategies. Table~\ref{tab:camera_strategy_dimension_gains} provides the corresponding changes on all seven perturbation dimensions for the completed Camera-view configurations. Each entry is the tuned or interpolated model's success rate minus that of the same original model, in percentage points. The Overall column pools all 10,030 episodes and is therefore weighted by the number of episodes in each dimension rather than being the arithmetic mean of the seven dimension-level gains. Aggregate values in the main text are computed before rounding, so the displayed one-decimal values can differ from a recomputed aggregate by $0.1$ point. For WiSE-FT, $\alpha=0$ is omitted as the original model with zero gain, and $\alpha=1$ is the Dual-LoRA row.

\begin{table}[h]
  \caption{Dimension-level gains under Camera-view tuning and retention strategies. All values are percentage points.}
  \label{tab:camera_strategy_dimension_gains}
  \centering
  \setlength{\tabcolsep}{3.5pt}
  \renewcommand{\arraystretch}{1.02}
  \begin{tabular*}{\linewidth}{@{\extracolsep{\fill}}lrrrrrrrr@{}}
    \toprule
    \textbf{Strategy} & \textbf{Cam.} & \textbf{Robot} & \textbf{Lang.} & \textbf{Light} & \textbf{Back.} & \textbf{Noise} & \textbf{Layout} & \textbf{Overall} \\
    \midrule
    \multicolumn{9}{l}{\textbf{X-VLA}} \\
    \cmidrule(lr){1-9}
    Connector only & $+36.5$ & $-4.1$ & $-13.3$ & $-6.0$ & $-5.2$ & $-9.6$ & $-4.0$ & $-0.2$ \\
    Backbone only & $+69.4$ & $-0.1$ & $-4.9$ & $-1.8$ & $-2.2$ & $-10.6$ & $-3.7$ & $+7.6$ \\
    Dual LoRA & $+69.4$ & $+0.5$ & $-6.6$ & $-2.2$ & $-3.9$ & $-10.7$ & $-4.9$ & $+7.0$ \\
    WiSE-FT $\alpha=0.25$ & $+11.6$ & $-0.1$ & $+0.3$ & $+1.7$ & $+0.9$ & $+0.7$ & $+2.0$ & $+2.6$ \\
    WiSE-FT $\alpha=0.50$ & $+32.0$ & $+1.4$ & $+1.5$ & $+0.8$ & $+2.0$ & $-0.6$ & $+1.7$ & $+6.0$ \\
    WiSE-FT $\alpha=0.75$ & $+59.8$ & $+1.9$ & $-0.5$ & $+0.4$ & $+0.9$ & $-3.4$ & $+0.6$ & $+9.4$ \\
    DELTA-style & $+62.7$ & $+2.6$ & $+1.2$ & $+1.0$ & $+2.8$ & $+0.8$ & $-1.2$ & $+10.9$ \\
    \addlinespace[2pt]
    \multicolumn{9}{l}{\textbf{OpenVLA-OFT}} \\
    \cmidrule(lr){1-9}
    Connector only & $+33.8$ & $+2.3$ & $-2.5$ & $+0.0$ & $+1.0$ & $+6.7$ & $+0.2$ & $+6.6$ \\
    Backbone only & $+37.8$ & $+0.4$ & $+0.1$ & $+1.2$ & $+0.3$ & $+8.2$ & $-3.0$ & $+7.1$ \\
    Dual LoRA & $+39.3$ & $+0.3$ & $+0.1$ & $+1.0$ & $+2.0$ & $+7.9$ & $-2.8$ & $+7.5$ \\
    WiSE-FT $\alpha=0.25$ & $+10.9$ & $+0.6$ & $+1.2$ & $-1.9$ & $+3.3$ & $+4.0$ & $+2.0$ & $+3.1$ \\
    WiSE-FT $\alpha=0.50$ & $+23.6$ & $+0.0$ & $+0.4$ & $+0.2$ & $+2.9$ & $+5.7$ & $+0.8$ & $+5.2$ \\
    WiSE-FT $\alpha=0.75$ & $+36.9$ & $+1.0$ & $+0.7$ & $+1.1$ & $+3.8$ & $+7.5$ & $-0.1$ & $+7.9$ \\
    DELTA-style & $+13.6$ & $-7.2$ & $-27.2$ & $-23.9$ & $-37.8$ & $-25.5$ & $-17.2$ & $-16.6$ \\
    \bottomrule
  \end{tabular*}
\end{table}

\section{Checkpoint and Training-Seed Analysis}
\label{app:checkpoint_analysis}

The main benchmark compares the final checkpoints on all 10,030 episodes. This section complements that comparison by examining intermediate checkpoints and independent training seeds on the same balanced 560-episode subset. Figure~\ref{fig:checkpoint_dynamics} and Table~\ref{tab:training_seed_sensitivity} test whether the model-dependent retention differences can be reduced to checkpoint selection or a single training run.

\subsection{Checkpoint Trajectories}

\begin{figure}[h]
  \centering
  \includegraphics[width=\linewidth]{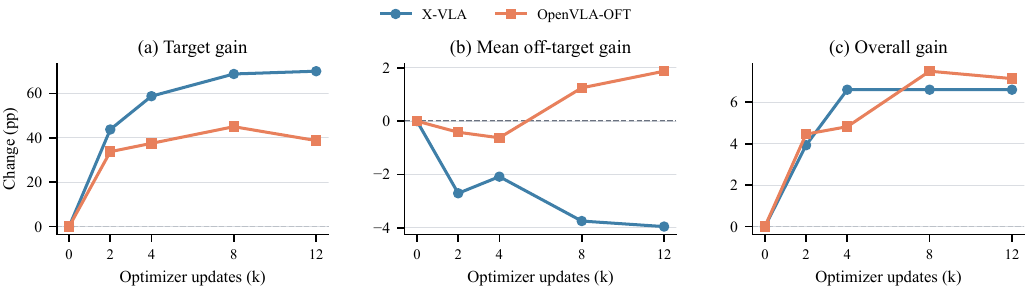}
  \caption{\textbf{Camera-view checkpoint dynamics.} Target, mean off-target, and overall gains are measured relative to the base model on a fixed balanced subset. All gains are reported in percentage points. The target and off-target changes do not evolve monotonically together.}
  \label{fig:checkpoint_dynamics}
\end{figure}

Figure~\ref{fig:checkpoint_dynamics} plots target, mean off-target, and overall gains at intermediate Camera-view Dual-LoRA checkpoints for X-VLA and OpenVLA-OFT. Every point is measured against the corresponding base model on the same subset of 80 episodes from each perturbation dimension. The curves show that target improvement and off-target regression do not evolve monotonically together. From 2k to 4k updates, X-VLA's target gain rises from $+43.8$ to $+58.8$ while its mean off-target gain improves from $-2.7$ to $-2.1$; it later reaches $+70.0$ target gain with $-4.0$ off target. OpenVLA-OFT reaches $+45.0$ and $+1.2$ at 8k, then $+38.8$ and $+1.9$ at 12k. Greater target improvement therefore does not imply greater off-target regression, and the two models stay separated throughout training.

\subsection{Training-Seed Sensitivity}
\label{app:training_seed_sensitivity}

We repeat Camera-view Dual-LoRA training with seeds 42, 43, and 44 for X-VLA and OpenVLA-OFT. Every final checkpoint is evaluated with the same evaluation seed on the balanced 560-episode subset used for Figure~\ref{fig:checkpoint_dynamics}. Table~\ref{tab:training_seed_sensitivity} reports target, mean off-target, and overall gains for each run. X-VLA has negative mean off-target gains of $-4.0$, $-5.0$, and $-7.1$ across the three seeds, whereas OpenVLA-OFT has positive gains of $+1.9$, $+3.8$, and $+1.2$. The model contrast is therefore present in every evaluated seed. To sum up, the checkpoint and seed analyses indicate that the model dependence in Section~\ref{sec:target_tuning} is not an artifact of one checkpoint or training run. This smaller sensitivity analysis complements, rather than replaces, the full 10,030-episode comparison in Table~\ref{tab:targeted_tuning_results}.

\begin{table}[h]
  \caption{Camera-view sensitivity to training seed on the fixed 560-episode subset. All values are percentage-point gains over the corresponding base model on the same episodes.}
  \label{tab:training_seed_sensitivity}
  \centering
  \setlength{\tabcolsep}{5pt}
  \renewcommand{\arraystretch}{1.02}
  \begin{tabular*}{\linewidth}{@{\extracolsep{\fill}}lrrrr@{}}
    \toprule
    \textbf{Model} & \textbf{Seed} & $\boldsymbol{G^{\mathrm{target}}}$ & $\boldsymbol{G^{\mathrm{off}}}$ & $\boldsymbol{G^{\mathrm{all}}}$ \\
    \midrule
    X-VLA & 42 & $+70.0$ & $-4.0$ & $+6.6$ \\
    X-VLA & 43 & $+73.8$ & $-5.0$ & $+6.3$ \\
    X-VLA & 44 & $+70.0$ & $-7.1$ & $+3.9$ \\
    OpenVLA-OFT & 42 & $+38.8$ & $+1.9$ & $+7.1$ \\
    OpenVLA-OFT & 43 & $+40.0$ & $+3.8$ & $+8.9$ \\
    OpenVLA-OFT & 44 & $+41.2$ & $+1.2$ & $+7.0$ \\
    \bottomrule
  \end{tabular*}
\end{table}

%% file: appendix_ci.tex
\section{Cluster-Bootstrap Confidence Intervals}
\label{app:confidence_intervals}

We compute pointwise $95\%$ confidence intervals using a paired cluster bootstrap over the 40 LIBERO task families, each comprising one source task and all its perturbed episodes. Every draw samples task families with replacement and retains all evaluation episodes and perturbation variants associated with each sampled family. The same draws are applied to the base and tuned results, and all metrics are recomputed before taking the 2.5th and 97.5th percentiles. These intervals quantify task-family sampling uncertainty conditional on the evaluated checkpoints; they do not estimate training-seed or repeated-rollout variation and are not adjusted for multiple comparisons.

Tables~\ref{tab:camera_ci_summary} and~\ref{tab:noise_ci_summary} report intervals for the four metrics in Table~\ref{tab:targeted_tuning_results}. The interval for $G^{\mathrm{worst}}$ is computed by taking the minimum off-target change within every bootstrap draw, rather than reusing the interval of the dimension with the lowest point estimate. The remaining tables report all dimension-level paired changes underlying Figure~\ref{fig:cross_dimension_changes}.

\begin{table}[h]
  \caption{Camera-view tuning: summary gains and pointwise $95\%$ cluster-bootstrap confidence intervals. Values are percentage points.}
  \label{tab:camera_ci_summary}
  \centering
  \setlength{\tabcolsep}{2pt}
  \renewcommand{\arraystretch}{1.08}
  \begin{tabular*}{\linewidth}{@{\extracolsep{\fill}}lcccc@{}}
    \toprule
    \textbf{Model} & $\boldsymbol{G^{\mathrm{target}}}$ & $\boldsymbol{G^{\mathrm{off}}}$ & $\boldsymbol{G^{\mathrm{worst}}}$ & $\boldsymbol{G^{\mathrm{all}}}$ \\
    \midrule
    OpenVLA-OFT & \shortstack{$+39.3$\\$[+30.6,+47.4]$} & \shortstack{$+1.4$\\$[+0.2,+2.8]$} & \shortstack{$-2.8$\\$[-5.1,-0.9]$} & \shortstack{$+7.5$\\$[+5.4,+9.6]$} \\
    $\pi_{0.5}$ & \shortstack{$+22.0$\\$[+15.2,+28.8]$} & \shortstack{$-0.3$\\$[-1.7,+1.2]$} & \shortstack{$-3.9$\\$[-6.3,-1.6]$} & \shortstack{$+3.0$\\$[+1.4,+4.7]$} \\
    InternVLA-A1.5 & \shortstack{$+2.6$\\$[-4.0,+9.2]$} & \shortstack{$-5.2$\\$[-7.3,-3.5]$} & \shortstack{$-12.4$\\$[-16.1,-8.9]$} & \shortstack{$-4.2$\\$[-6.5,-2.1]$} \\
    NORA-1.5 & \shortstack{$+28.7$\\$[+22.3,+35.4]$} & \shortstack{$-10.1$\\$[-13.9,-6.7]$} & \shortstack{$-17.8$\\$[-26.6,-11.3]$} & \shortstack{$-3.7$\\$[-7.2,-0.4]$} \\
    X-VLA & \shortstack{$+69.4$\\$[+61.3,+76.4]$} & \shortstack{$-4.6$\\$[-7.4,-2.1]$} & \shortstack{$-10.7$\\$[-15.1,-7.1]$} & \shortstack{$+7.0$\\$[+4.3,+9.6]$} \\
    \bottomrule
  \end{tabular*}
\end{table}

\begin{table}[h]
  \caption{Sensor-noise tuning: summary gains and pointwise $95\%$ cluster-bootstrap confidence intervals. Values are percentage points.}
  \label{tab:noise_ci_summary}
  \centering
  \setlength{\tabcolsep}{2pt}
  \renewcommand{\arraystretch}{1.08}
  \begin{tabular*}{\linewidth}{@{\extracolsep{\fill}}lcccc@{}}
    \toprule
    \textbf{Model} & $\boldsymbol{G^{\mathrm{target}}}$ & $\boldsymbol{G^{\mathrm{off}}}$ & $\boldsymbol{G^{\mathrm{worst}}}$ & $\boldsymbol{G^{\mathrm{all}}}$ \\
    \midrule
    OpenVLA-OFT & \shortstack{$+19.8$\\$[+13.4,+26.7]$} & \shortstack{$+0.3$\\$[-1.1,+1.7]$} & \shortstack{$-0.6$\\$[-5.0,+0.3]$} & \shortstack{$+3.4$\\$[+2.0,+4.9]$} \\
    $\pi_{0.5}$ & \shortstack{$+10.2$\\$[+6.7,+14.0]$} & \shortstack{$-1.3$\\$[-2.1,-0.7]$} & \shortstack{$-4.1$\\$[-6.2,-3.0]$} & \shortstack{$+0.3$\\$[-0.4,+1.1]$} \\
    InternVLA-A1.5 & \shortstack{$-1.9$\\$[-6.4,+2.3]$} & \shortstack{$-2.5$\\$[-3.4,-1.6]$} & \shortstack{$-6.0$\\$[-8.9,-3.7]$} & \shortstack{$-2.6$\\$[-3.7,-1.5]$} \\
    NORA-1.5 & \shortstack{$+45.3$\\$[+38.6,+51.6]$} & \shortstack{$-1.0$\\$[-3.4,+1.4]$} & \shortstack{$-4.4$\\$[-9.4,-2.6]$} & \shortstack{$+6.2$\\$[+3.7,+8.6]$} \\
    X-VLA & \shortstack{$+32.8$\\$[+27.5,+38.0]$} & \shortstack{$-2.2$\\$[-3.8,-0.7]$} & \shortstack{$-4.2$\\$[-8.1,-2.6]$} & \shortstack{$+3.2$\\$[+1.6,+4.9]$} \\
    \bottomrule
  \end{tabular*}
\end{table}

\subsection{Dimension-Level Paired Changes}

Tables~\ref{tab:ci_openvla}--\ref{tab:ci_xvla} decompose the summary metrics into the seven paired dimension-level changes for each model. Rows identify the evaluation perturbation, while the two columns identify whether the visual pathway was tuned on Camera-view or Sensor-noise data; the final row gives the episode-weighted overall change. Each cell reports the point estimate followed by its pointwise $95\%$ cluster-bootstrap confidence interval. These tables show which individual regressions underlie the mean and worst off-target summaries in Table~\ref{tab:targeted_tuning_results}. In particular, intervals below zero occur across language, robot-state, object-layout, and visual perturbations, showing that the measured off-target effects are distributed across multiple evaluation conditions rather than driven by a single dimension.

\begin{table}[h]
  \caption{OpenVLA-OFT: paired dimension-level changes and pointwise $95\%$ cluster-bootstrap confidence intervals. Values are percentage points.}
  \label{tab:ci_openvla}
  \centering
  \setlength{\tabcolsep}{8pt}
  \renewcommand{\arraystretch}{1.04}
  \begin{tabular*}{\linewidth}{@{\extracolsep{\fill}}lrr@{}}
    \toprule
    \textbf{Evaluation dimension} & \textbf{Camera-view tuning} & \textbf{Sensor-noise tuning} \\
    \midrule
    Camera view & $+39.3\;[+30.6,\,+47.4]$ & $+0.0\;[-4.3,\,+4.4]$ \\
    Robot state & $+0.3\;[-1.4,\,+2.1]$ & $-0.1\;[-1.5,\,+1.3]$ \\
    Language & $+0.1\;[-1.4,\,+1.7]$ & $-0.2\;[-1.6,\,+1.2]$ \\
    Lighting & $+1.0\;[-2.0,\,+3.9]$ & $-0.6\;[-4.5,\,+2.6]$ \\
    Background & $+2.0\;[-2.5,\,+6.4]$ & $+1.7\;[-1.0,\,+4.4]$ \\
    Sensor noise & $+7.9\;[+4.7,\,+11.4]$ & $+19.8\;[+13.4,\,+26.7]$ \\
    Object layout & $-2.8\;[-5.0,\,-0.7]$ & $+1.1\;[-1.1,\,+3.7]$ \\
    Overall & $+7.5\;[+5.4,\,+9.6]$ & $+3.4\;[+2.0,\,+4.9]$ \\
    \bottomrule
  \end{tabular*}
\end{table}

\begin{table}[h]
  \caption{$\pi_{0.5}$: paired dimension-level changes and pointwise $95\%$ cluster-bootstrap confidence intervals. Values are percentage points.}
  \label{tab:ci_pi05}
  \centering
  \setlength{\tabcolsep}{8pt}
  \renewcommand{\arraystretch}{1.04}
  \begin{tabular*}{\linewidth}{@{\extracolsep{\fill}}lrr@{}}
    \toprule
    \textbf{Evaluation dimension} & \textbf{Camera-view tuning} & \textbf{Sensor-noise tuning} \\
    \midrule
    Camera view & $+22.0\;[+15.2,\,+28.8]$ & $-4.1\;[-6.0,\,-2.1]$ \\
    Robot state & $-0.5\;[-3.1,\,+2.3]$ & $-3.5\;[-5.9,\,-1.2]$ \\
    Language & $-1.1\;[-3.5,\,+1.5]$ & $-1.1\;[-2.7,\,+0.4]$ \\
    Lighting & $+3.0\;[+0.4,\,+6.0]$ & $-0.1\;[-2.1,\,+1.8]$ \\
    Background & $+1.6\;[-1.0,\,+4.4]$ & $+0.9\;[-0.3,\,+2.3]$ \\
    Sensor noise & $-3.9\;[-6.3,\,-1.4]$ & $+10.2\;[+6.7,\,+14.0]$ \\
    Object layout & $-1.0\;[-2.4,\,+0.4]$ & $-0.1\;[-1.4,\,+1.1]$ \\
    Overall & $+3.0\;[+1.4,\,+4.7]$ & $+0.3\;[-0.4,\,+1.1]$ \\
    \bottomrule
  \end{tabular*}
\end{table}

\begin{table}[h]
  \caption{InternVLA-A1.5: paired dimension-level changes and pointwise $95\%$ cluster-bootstrap confidence intervals. Values are percentage points.}
  \label{tab:ci_internvla}
  \centering
  \setlength{\tabcolsep}{8pt}
  \renewcommand{\arraystretch}{1.04}
  \begin{tabular*}{\linewidth}{@{\extracolsep{\fill}}lrr@{}}
    \toprule
    \textbf{Evaluation dimension} & \textbf{Camera-view tuning} & \textbf{Sensor-noise tuning} \\
    \midrule
    Camera view & $+2.6\;[-4.0,\,+9.2]$ & $-6.0\;[-8.9,\,-3.2]$ \\
    Robot state & $-12.4\;[-16.1,\,-8.9]$ & $-3.4\;[-5.3,\,-1.5]$ \\
    Language & $-5.8\;[-9.1,\,-2.8]$ & $-2.8\;[-4.9,\,-0.8]$ \\
    Lighting & $-1.4\;[-3.8,\,+0.9]$ & $+0.7\;[-2.5,\,+5.2]$ \\
    Background & $-3.4\;[-7.5,\,-0.3]$ & $-3.0\;[-5.9,\,-0.6]$ \\
    Sensor noise & $-6.1\;[-11.6,\,-1.8]$ & $-1.9\;[-6.4,\,+2.3]$ \\
    Object layout & $-2.2\;[-4.0,\,-0.5]$ & $-0.7\;[-2.3,\,+1.0]$ \\
    Overall & $-4.2\;[-6.5,\,-2.1]$ & $-2.6\;[-3.7,\,-1.5]$ \\
    \bottomrule
  \end{tabular*}
\end{table}

\begin{table}[h]
  \caption{NORA-1.5: paired dimension-level changes and pointwise $95\%$ cluster-bootstrap confidence intervals. Values are percentage points.}
  \label{tab:ci_nora}
  \centering
  \setlength{\tabcolsep}{8pt}
  \renewcommand{\arraystretch}{1.04}
  \begin{tabular*}{\linewidth}{@{\extracolsep{\fill}}lrr@{}}
    \toprule
    \textbf{Evaluation dimension} & \textbf{Camera-view tuning} & \textbf{Sensor-noise tuning} \\
    \midrule
    Camera view & $+28.7\;[+22.3,\,+35.4]$ & $+0.9\;[-0.4,\,+2.2]$ \\
    Robot state & $-6.6\;[-11.4,\,-2.0]$ & $-4.4\;[-6.9,\,-1.9]$ \\
    Language & $-11.9\;[-18.6,\,-5.4]$ & $-3.4\;[-6.6,\,-0.3]$ \\
    Lighting & $-7.5\;[-18.2,\,+3.3]$ & $+5.3\;[-2.3,\,+13.4]$ \\
    Background & $-17.8\;[-26.6,\,-9.6]$ & $-2.9\;[-9.4,\,+3.3]$ \\
    Sensor noise & $-5.9\;[-7.2,\,-4.6]$ & $+45.3\;[+38.6,\,+51.6]$ \\
    Object layout & $-11.1\;[-16.8,\,-5.9]$ & $-1.6\;[-3.7,\,+0.3]$ \\
    Overall & $-3.7\;[-7.2,\,-0.4]$ & $+6.2\;[+3.7,\,+8.6]$ \\
    \bottomrule
  \end{tabular*}
\end{table}

\begin{table}[h]
  \caption{X-VLA: paired dimension-level changes and pointwise $95\%$ cluster-bootstrap confidence intervals. Values are percentage points.}
  \label{tab:ci_xvla}
  \centering
  \setlength{\tabcolsep}{8pt}
  \renewcommand{\arraystretch}{1.04}
  \begin{tabular*}{\linewidth}{@{\extracolsep{\fill}}lrr@{}}
    \toprule
    \textbf{Evaluation dimension} & \textbf{Camera-view tuning} & \textbf{Sensor-noise tuning} \\
    \midrule
    Camera view & $+69.4\;[+61.3,\,+76.4]$ & $-2.8\;[-5.1,\,-0.7]$ \\
    Robot state & $+0.5\;[-3.2,\,+4.1]$ & $-3.3\;[-6.1,\,-0.5]$ \\
    Language & $-6.6\;[-10.1,\,-3.2]$ & $-4.2\;[-7.9,\,-0.8]$ \\
    Lighting & $-2.2\;[-8.2,\,+3.7]$ & $+1.2\;[-1.7,\,+5.9]$ \\
    Background & $-3.9\;[-9.0,\,+1.0]$ & $-2.2\;[-6.5,\,+1.7]$ \\
    Sensor noise & $-10.7\;[-15.1,\,-6.7]$ & $+32.8\;[+27.5,\,+38.0]$ \\
    Object layout & $-4.9\;[-9.2,\,-0.6]$ & $-2.0\;[-4.0,\,+0.1]$ \\
    Overall & $+7.0\;[+4.3,\,+9.6]$ & $+3.2\;[+1.6,\,+4.9]$ \\
    \bottomrule
  \end{tabular*}
\end{table}

\clearpage

%% file: iclr2027_conference.bib
@InProceedings{brohan2023rt2,
  title = 	 {{RT-2}: Vision-Language-Action Models Transfer Web Knowledge to Robotic Control},
  author =       {Zitkovich, Brianna and Yu, Tianhe and Xu, Sichun and Xu, Peng and Xiao, Ted and Xia, Fei and Wu, Jialin and Wohlhart, Paul and Welker, Stefan and Wahid, Ayzaan and Vuong, Quan and Vanhoucke, Vincent and Tran, Huong and Soricut, Radu and Singh, Anikait and Singh, Jaspiar and Sermanet, Pierre and Sanketi, Pannag R. and Salazar, Grecia and Ryoo, Michael S. and Reymann, Krista and Rao, Kanishka and Pertsch, Karl and Mordatch, Igor and Michalewski, Henryk and Lu, Yao and Levine, Sergey and Lee, Lisa and Lee, Tsang-Wei Edward and Leal, Isabel and Kuang, Yuheng and Kalashnikov, Dmitry and Julian, Ryan and Joshi, Nikhil J. and Irpan, Alex and Ichter, Brian and Hsu, Jasmine and Herzog, Alexander and Hausman, Karol and Gopalakrishnan, Keerthana and Fu, Chuyuan and Florence, Pete and Finn, Chelsea and Dubey, Kumar Avinava and Driess, Danny and Ding, Tianli and Choromanski, Krzysztof Marcin and Chen, Xi and Chebotar, Yevgen and Carbajal, Justice and Brown, Noah and Brohan, Anthony and Arenas, Montserrat Gonzalez and Han, Kehang},
  booktitle = 	 {Proceedings of The 7th Conference on Robot Learning},
  pages = 	 {2165--2183},
  year = 	 {2023},
  editor = 	 {Tan, Jie and Toussaint, Marc and Darvish, Kourosh},
  volume = 	 {229},
  series = 	 {Proceedings of Machine Learning Research},
  month = 	 {06--09 Nov},
  publisher =    {PMLR},
  url = 	 {https://proceedings.mlr.press/v229/zitkovich23a.html}
}

@inproceedings{driess2023palme,
author = {Driess, Danny and Xia, Fei and Sajjadi, Mehdi S. M. and Lynch, Corey and Chowdhery, Aakanksha and Ichter, Brian and Wahid, Ayzaan and Tompson, Jonathan and Vuong, Quan and Yu, Tianhe and Huang, Wenlong and Chebotar, Yevgen and Sermanet, Pierre and Duckworth, Daniel and Levine, Sergey and Vanhoucke, Vincent and Hausman, Karol and Toussaint, Marc and Greff, Klaus and Zeng, Andy and Mordatch, Igor and Florence, Pete},
title = {{PaLM-E}: an embodied multimodal language model},
year = {2023},
publisher = {JMLR.org},
booktitle = {Proceedings of the 40th International Conference on Machine Learning},
articleno = {340},
numpages = {20},
location = {Honolulu, Hawaii, USA},
series = {ICML'23}
}

@INPROCEEDINGS{openx2023,
  author={Open X-Embodiment Collaboration},
  booktitle={2024 IEEE International Conference on Robotics and Automation (ICRA)}, 
  title={Open {X-Embodiment}: Robotic Learning Datasets and {RT-X} Models}, 
  year={2024},
  volume={},
  number={},
  pages={6892-6903},
  doi={10.1109/ICRA57147.2024.10611477}}

@InProceedings{kim2024openvla,
  title = 	 {{OpenVLA}: An Open-Source Vision-Language-Action Model},
  author =       {Kim, Moo Jin and Pertsch, Karl and Karamcheti, Siddharth and Xiao, Ted and Balakrishna, Ashwin and Nair, Suraj and Rafailov, Rafael and Foster, Ethan P and Sanketi, Pannag R and Vuong, Quan and Kollar, Thomas and Burchfiel, Benjamin and Tedrake, Russ and Sadigh, Dorsa and Levine, Sergey and Liang, Percy and Finn, Chelsea},
  booktitle = 	 {Proceedings of The 8th Conference on Robot Learning},
  pages = 	 {2679--2713},
  year = 	 {2025},
  editor = 	 {Agrawal, Pulkit and Kroemer, Oliver and Burgard, Wolfram},
  volume = 	 {270},
  series = 	 {Proceedings of Machine Learning Research},
  month = 	 {06--09 Nov},
  publisher =    {PMLR},
  url = 	 {https://proceedings.mlr.press/v270/kim25c.html}
}

@INPROCEEDINGS{kim2025oft, 
    AUTHOR    = {Moo Jin Kim AND Chelsea Finn AND Percy Liang}, 
    TITLE     = {{Fine-Tuning Vision-Language-Action Models: Optimizing Speed and Success}}, 
    BOOKTITLE = {Proceedings of Robotics: Science and Systems}, 
    YEAR      = {2025}, 
    ADDRESS   = {LosAngeles, CA, USA}, 
    MONTH     = {June}, 
    DOI       = {10.15607/RSS.2025.XXI.017} 
}

@InProceedings{black2025pi05,
  title = 	 {$\pi_{0.5}$: a Vision-Language-Action Model with Open-World Generalization},
  author =       {Black, Kevin and Brown, Noah and Darpinian, James and Dhabalia, Karan and Driess, Danny and Esmail, Adnan and Equi, Michael Robert and Finn, Chelsea and Fusai, Niccolo and Galliker, Manuel Y. and Ghosh, Dibya and Groom, Lachy and Hausman, Karol and ichter, brian and Jakubczak, Szymon and Jones, Tim and Ke, Liyiming and LeBlanc, Devin and Levine, Sergey and Li-Bell, Adrian and Mothukuri, Mohith and Nair, Suraj and Pertsch, Karl and Ren, Allen Z. and Shi, Lucy Xiaoyang and Smith, Laura and Springenberg, Jost Tobias and Stachowicz, Kyle and Tanner, James and Vuong, Quan and Walke, Homer and Walling, Anna and Wang, Haohuan and Yu, Lili and Zhilinsky, Ury},
  booktitle = 	 {Proceedings of The 9th Conference on Robot Learning},
  pages = 	 {17--40},
  year = 	 {2025},
  editor = 	 {Lim, Joseph and Song, Shuran and Park, Hae-Won},
  volume = 	 {305},
  series = 	 {Proceedings of Machine Learning Research},
  month = 	 {27--30 Sep},
  publisher =    {PMLR},
  url = 	 {https://proceedings.mlr.press/v305/black25a.html}
}

@inproceedings{
zheng2025xvla,
title={{X-VLA}: Soft-Prompted {Transformer} as Scalable Cross-Embodiment Vision-Language-Action Model},
author={Jinliang Zheng and Jianxiong Li and Zhihao Wang and Dongxiu Liu and Xirui Kang and Yuchun Feng and Yinan Zheng and Jiayin Zou and Yilun Chen and Jia Zeng and Tai Wang and Ya-Qin Zhang and Jingjing Liu and Xianyuan Zhan},
booktitle={The Fourteenth International Conference on Learning Representations},
year={2026},
url={https://openreview.net/forum?id=kt51kZH4aG}
}

@misc{hung2025nora,
      title={{NORA-1.5}: A Vision-Language-Action Model Trained using World Model- and Action-based Preference Rewards}, 
      author={Chia-Yu Hung and Navonil Majumder and Haoyuan Deng and Liu Renhang and Yankang Ang and Amir Zadeh and Chuan Li and Dorien Herremans and Ziwei Wang and Soujanya Poria},
      year={2025},
      eprint={2511.14659},
      archivePrefix={arXiv},
      primaryClass={cs.RO},
      url={https://arxiv.org/abs/2511.14659}, 
}

@misc{ma2026internvla,
      title={{InternVLA-A1.5}: Unifying Understanding, Latent Foresight, and Action for Compositional Generalization}, 
      author={Haoxiang Ma and Junhao Cai and Xiaoxu Xu and Hao Li and Yuyin Yang and Yang Tian and Jiafei Cao and Hongrui Zhu and Zherui Qiu and Zhaxizhuoma and Yuqiang Yang and Jiaqi Peng and Xueyuan Wei and Yangkun Zhu and Jiahao Jiang and Xing Gao and Hanqing Wang and Feng Yuan and Kailin Li and Xueyue Zhu and Tai Wang and Yan Ding and Jiangmiao Pang and Jia Zeng and Jingjing Zhang and Bowen Zhou and Yao Mu and Chunhua Shen and Weinan Zhang},
      year={2026},
      eprint={2607.04988},
      archivePrefix={arXiv},
      primaryClass={cs.RO},
      url={https://arxiv.org/abs/2607.04988}, 
}

@misc{simplerenv2024,
      title={Evaluating Real-World Robot Manipulation Policies in Simulation}, 
      author={Xuanlin Li and Kyle Hsu and Jiayuan Gu and Karl Pertsch and Oier Mees and Homer Rich Walke and Chuyuan Fu and Ishikaa Lunawat and Isabel Sieh and Sean Kirmani and Sergey Levine and Jiajun Wu and Chelsea Finn and Hao Su and Quan Vuong and Ted Xiao},
      year={2024},
      eprint={2405.05941},
      archivePrefix={arXiv},
      primaryClass={cs.RO},
      url={https://arxiv.org/abs/2405.05941}, 
}

@InProceedings{fei2025liberoplus,
    author    = {Fei, Senyu and Wang, Siyin and Shi, Junhao and Dai, Zihao and Cai, Jikun and Qian, Pengfang and Ji, Li and He, Xinzhe and Zhang, Shiduo and Fei, Zhaoye and Fu, Jinlan and Gong, Jingjing and Qiu, Xipeng},
    title     = {{LIBERO-Plus}: A Progressive Robustness Benchmark for Visual-Language-Action Models},
    booktitle = {Proceedings of the IEEE/CVF Conference on Computer Vision and Pattern Recognition (CVPR)},
    month     = {June},
    year      = {2026},
    pages     = {38574-38583}
}

@misc{li2025generalizable,
      title={{VLA} Models Are More Generalizable Than You Think: Revisiting Physical and Spatial Modeling}, 
      author={Weiqi Li and Quande Zhang and Ruifeng Zhai and Liang Lin and Guangrun Wang},
      year={2026},
      eprint={2512.02902},
      archivePrefix={arXiv},
      primaryClass={cs.RO},
      url={https://arxiv.org/abs/2512.02902}, 
}

@misc{zhang2026sarf,
      title={Structure-Aware Robust Fine-Tuning: Defending Vision-Language-Action Robots Against Physical Attention Hijacking}, 
      author={Jinquan Zhang and Dongfu Yin and Run Yang and Yufeng Yan and Zhen Tian and F. Richard Yu},
      year={2026},
      eprint={2608.03231},
      archivePrefix={arXiv},
      primaryClass={cs.RO},
      url={https://arxiv.org/abs/2608.03231}, 
}

@misc{yin2026vlaguard,
      title={{VLAGuard}: A Framework for Evaluating and Mitigating Physical Attention Hijacking in Vision-Language-Action Robots within Wireless Sensor Networks}, 
      author={Dongfu Yin and Jinquan Zhang},
      year={2026},
      eprint={2608.01028},
      archivePrefix={arXiv},
      primaryClass={cs.RO},
      url={https://arxiv.org/abs/2608.01028}, 
}

@inproceedings{
hu2022lora,
title={Lo{RA}: Low-Rank Adaptation of Large Language Models},
author={Edward J Hu and yelong shen and Phillip Wallis and Zeyuan Allen-Zhu and Yuanzhi Li and Shean Wang and Lu Wang and Weizhu Chen},
booktitle={International Conference on Learning Representations},
year={2022},
url={https://openreview.net/forum?id=nZeVKeeFYf9}
}

@inproceedings{zhang2026vlm4vla,
  title={{VLM4VLA}: Revisiting Vision-Language Models in Vision-Language-Action Models},
  author={Zhang, Jianke and Chen, Xiaoyu and Wang, Qiuyue and Li, Mingsheng and Guo, Yanjiang and Hu, Yucheng and Zhang, Jiajun and Bai, Shuai and Lin, Junyang and Chen, Jianyu},
  booktitle={International Conference on Learning Representations},
  year={2026},
  url={https://arxiv.org/abs/2601.03309}
}

@inproceedings{
zhang2025vlaarena,
title={{VLA}-Arena: An Open-Source Framework for Benchmarking Vision-Language-Action Models},
author={Borong Zhang and Jiahao Li and Jiachen Shen and Yuhao Zhang and Yishuai Cai and Yuanpei Chen and Juntao Dai and Jiaming Ji and Yaodong Yang},
booktitle={Forty-third International Conference on Machine Learning},
year={2026},
url={https://openreview.net/forum?id=qjeCH95rbF}
}

@inproceedings{wang2026liberox,
  title={{LIBERO-X}: Robustness Litmus for Vision-Language-Action Models},
  author={Wang, Guodong and Zhang, Chenkai and Liu, Qingjie and Zhang, Jinjin and Cai, Jiancheng and Liu, Junjie and Liu, Xinmin},
  booktitle={Proceedings of Robotics: Science and Systems},
  address={Sydney, Australia},
  month={July},
  year={2026}
}

@inproceedings{
liu2023libero,
title={{LIBERO}: Benchmarking Knowledge Transfer for Lifelong Robot Learning},
author={Bo Liu and Yifeng Zhu and Chongkai Gao and Yihao Feng and qiang liu and Yuke Zhu and Peter Stone},
booktitle={Thirty-seventh Conference on Neural Information Processing Systems Datasets and Benchmarks Track},
year={2023},
url={https://openreview.net/forum?id=xzEtNSuDJk}
}

@article{choi2026vlaeval,
  title={{vla-eval}: A Unified Evaluation Harness for Vision-Language-Action Models},
  author={Choi, Suhwan and Lee, Yunsung and Park, Yubeen and Kim, Chris Dongjoo and Krishna, Ranjay and Fox, Dieter and Yu, Youngjae},
  journal={arXiv preprint arXiv:2603.13966},
  year={2026},
  url={https://arxiv.org/abs/2603.13966}
}

@inproceedings{loshchilov2019decoupled,
  title={Decoupled Weight Decay Regularization},
  author={Loshchilov, Ilya and Hutter, Frank},
  booktitle={International Conference on Learning Representations},
  year={2019},
  url={https://openreview.net/forum?id=Bkg6RiCqY7}
}

@InProceedings{wortsman2022robust,
  author    = {Wortsman, Mitchell and Ilharco, Gabriel and Kim, Jong Wook and Li, Mike and Kornblith, Simon and Roelofs, Rebecca and Lopes, Raphael Gontijo and Hajishirzi, Hannaneh and Farhadi, Ali and Namkoong, Hongseok and Schmidt, Ludwig},
  title     = {Robust Fine-Tuning of Zero-Shot Models},
  booktitle = {Proceedings of the IEEE/CVF Conference on Computer Vision and Pattern Recognition (CVPR)},
  year      = {2022},
  pages     = {7959--7971}
}

@inproceedings{
li2018delta,
title={{DELTA}: {DEEP} {LEARNING} {TRANSFER} {USING} {FEATURE} {MAP} {WITH} {ATTENTION} {FOR} {CONVOLUTIONAL} {NETWORKS}},
author={Xingjian Li and Haoyi Xiong and Hanchao Wang and Yuxuan Rao and Liping Liu and Jun Huan},
booktitle={International Conference on Learning Representations},
year={2019},
url={https://openreview.net/forum?id=rkgbwsAcYm},
}

@inproceedings{todorov2012mujoco,
  title={MuJoCo: A physics engine for model-based control},
  author={Todorov, Emanuel and Erez, Tom and Tassa, Yuval},
  booktitle={2012 IEEE/RSJ International Conference on Intelligent Robots and Systems},
  pages={5026--5033},
  year={2012},
  organization={IEEE},
  doi={10.1109/IROS.2012.6386109}
}

@inproceedings{robosuite2020,
  title={robosuite: A Modular Simulation Framework and Benchmark for Robot Learning},
  author={Yuke Zhu and Josiah Wong and Ajay Mandlekar and Roberto Mart\'{i}n-Mart\'{i}n and Abhishek Joshi and Soroush Nasiriany and Yifeng Zhu and Kevin Lin},
  booktitle={arXiv preprint arXiv:2009.12293},
  year={2020}
}
